%% file: ijcai22.tex
\documentclass{article}
\usepackage{ijcai22}
	
\usepackage[lined,boxed,commentsnumbered,linesnumbered,ruled]{algorithm2e}
\usepackage{times}
\usepackage{multirow}
\usepackage{soul}
\usepackage{url}
\usepackage{stmaryrd}
\usepackage[hidelinks]{hyperref}
\usepackage[utf8]{inputenc}
\usepackage[small]{caption}
\usepackage{graphicx}
\usepackage{amsmath}
\usepackage{amsthm}
\usepackage{booktabs}

\newtheorem{problem}{Problem}
\usepackage{booktabs}
\usepackage{amssymb}
\usepackage{xspace}
\newcommand{\method}{\texttt{GOCPT}\xspace}
\newcommand{\old}{\mathrm{old}}

\newcommand{\diag}{\mathrm{diag}}
\newcommand{\init}{\mathrm{init}}

\newcommand{\update}{\mathrm{update}}
\newcommand{\reg}{\mathrm{reg}}
\newcommand{\rec}{\mathrm{rec}}

\newcommand{\light}{\mathrm{E}}

\usepackage{color}
\usepackage{xcolor}

\usepackage[normalem]{ulem} % \sout
\usepackage{enumitem}
\setlist[itemize]{leftmargin=*}
\usepackage{stackengine} 
\usepackage{wasysym}
\newcommand\oast{\stackMath\mathbin{\stackinset{c}{0ex}{c}{0ex}{\ast}{\ocircle}}}

\title{\method: Generalized Online Canonical  Polyadic  Tensor Factorization \\ and Completion}

\author{
}

\author{
Chaoqi Yang$^1$\and
Cheng Qian$^2$\And
Jimeng Sun$^{1*}$\\
\affiliations
$^1$Department of Computer Science, University of Illinois Urbana-Champaign \\ $^2$Analytics Center of Excellence, IQVIA
\emails
$^1$\{chaoqiy2, jimeng\}@illinois.edu,
$^2$alextoqc@gmail.com
}
\begin{document}

\maketitle

\begin{abstract}
  Low-rank tensor factorization or completion is well-studied and applied in various online settings, such as online tensor factorization (where the temporal mode grows) and online tensor completion (where incomplete slices arrive gradually). However, in many real-world settings, tensors may have more complex evolving patterns: (i) one or more modes can grow; (ii) missing entries may be filled; (iii) existing tensor elements can change. Existing methods cannot support such complex scenarios. To fill the gap, this paper proposes a \underline{G}eneralized \underline{O}nline \underline{C}anonical \underline{P}olyadic (CP) \underline{T}ensor factorization and completion framework (named \method) for this general setting, where we maintain the CP structure of such dynamic tensors during the evolution. We show that existing online tensor factorization and completion setups can be unified under the \method framework. Furthermore, 
we propose a variant, named $\mbox{\method}_{\light}$, to deal with cases where historical tensor elements are unavailable (e.g., privacy protection), which
achieves similar fitness as \method but with much less computational cost.
% \js{We should mention the existing online tensor methods first then state how we unify/generalize them.}
Experimental results
demonstrate that our \method can improve fitness by up to $2.8\%$ on the JHU Covid data and $9.2\%$ on a proprietary patient claim dataset over baselines. Our variant $\mbox{\method}_{\light}$ shows up to $1.2\%$ and $5.5\%$ fitness improvement on two datasets with about $20\%$ speedup compared to the best model.
\end{abstract}

\input{sec1-introduction}

\input{sec2-problem}
\input{sec3-method}

\input{sec4-unifying}

\input{sec5-experiment}

\input{sec6-conclusion}

% \vspace{-2mm}
\section*{Acknowledgements}
\vspace{-1mm}
This work was in part supported by the National Science Foundation award SCH-2014438, PPoSS 2028839, IIS-1838042, the National Institute of Health award NIH R01 1R01NS107291-01 and OSF Healthcare. We thank Navjot Singh and Prof. Edgar Solomonik for valuable discussions.

\clearpage

\bibliographystyle{named}
\bibliography{ijcai22}
\clearpage
\appendix
\input{appendix}

\end{document}

%% file: sec1-introduction.tex
\section{Introduction}
%Multidimensional data are usually incomplete due to unpredictable reasons, such as limited permission, and unavoidable misconduct \cite{song2019tensor}. To impute the missing values, previous works model the data structures as tensors and propose \cite{yang2021mtc} tensor completion methods to simultaneously fill the missings and capture the low-rank data components.  Under the most well-known Canonical Polyadic (CP) structure, two popular methods for handling tensor completion (CPC) are (i) imputing the tensor and then conducting standard CP decomposition iteratively \cite{yang2021mtc,qian2021multi}; (ii) directly dealing with the derivative of the masked objective \cite{mardani2014imputation,mardani2015subspace}.

Streaming tensor data becomes increasingly popular in areas such as spatio-temporal outlier detection \cite{najafi2019outlier}, social media \cite{song2017multi}, sensor monitoring \cite{mardani2015subspace}, video analysis \cite{kasai2019fast} and hyper-order time series \cite{cai2015facets}. 
The factorization/decomposition of such multidimensional structural data is challenging since they are usually sparse, delayed, and sometimes incomplete.
% online tensor factorization and completion have benefited various applications, such as filling the missing value \cite{zheng2021fully} and data compression \cite{gujral2019octen}. 
% The streaming data is usually sparse, delayed and sometimes incomplete, while the factorization/decomposition of such multidimensional structural data is challenging. 
There is an increasing need to maintain the low-rank Tucker \cite{sun2006beyond,xiao2018eotd,nimishakavi2018inductive,fang2021bayesian,gilman2020grassmannian} or CP \cite{du2018probabilistic,phipps2021streaming} structure
of the evolving tensors
in such dynamics, considering model efficiency and scalability. 

Several online (we also use ``streaming" interchangeably) settings have been discussed before. The most popular two are {\em online tensor decomposition} \cite{zhou2016accelerating,song2017multi} and {\em online tensor completion} \cite{kasai2019fast}, where the temporal mode grows with new incoming slices.
% \edgar{The following paragraph on related work would be more useful after terminology and background on the problem is introduced, it is better to give a high-level summary / mention only key works in intro.}
Some pioneer works have been proposed for these two particular settings. For the factorization problem,
\cite{nion2009adaptive} incrementally tracked the singular value decomposition (SVD) of the unfolded third-order tensors to maintain the CP factorization results. Accelerated methods have been proposed for the evolving dense \cite{zhou2016accelerating} or sparse \cite{zhou2018online} tensors by reusing intermediate quantities, e.g., matricized tensor times Khatri-Rao product (MTTKRP). For the completion problem, recursive least squares \cite{kasai2016online} and stochastic gradient descent (SGD) \cite{mardani2015subspace} were studied to track the evolving subspace of incomplete data.

% \edgar{What is this an example of? How does our work compare to the above lines of work specifically?}
However, most existing methods are designed for the growing patterns. They cannot support other possible patterns (e.g., missing entries in the previous tensor can be refilled or existing values can be updated) or more complex scenarios where the tensors evolve with a combination of patterns.%\sout{, or new tensor slices can be added.} 

\smallskip
\noindent{\bf Motivating application: } Let us consider a public health surveillance application where data is modeled as a fourth-order tensor indexed by {\em location} (e.g., zip code), {\em disease} (e.g., diagnosis code), {\em data generation date (GD)} (i.e., the date when the clinical events actually occur) and {\em data loading date (LD)} (i.e., the time when the events are reported in the database) and each tensor element stands for the count of medical claims with particular location-disease-date tuples. 
The tensor grows over time by adding new locations, diseases, GD's, or LD's. For every new LD, missing entries before that date may be filled in, and existing entries can be revised for data correction purposes \cite{qian2021multi}.
%Every week, this tensor grows (along with the third mode) since new service will be provided; missing entries are filled because the insurance reports can delay by few weeks; existing tensor entries can change due to unpredictable reasons, such as data correction. 
% Similar settings can appear in online display advertising, where the backflow data from the advertisers are usually delayed. 
%Maintaining a low-rank factorization/completion result of such complex data is important, but few works have studied this setting in the literature.
Dealing with such a dynamic tensor is challenging, and very few works have studied this online tensor factorization/completion problem. The most recent work in  \cite{qian2021multi} can only handle a special case with the GD dimension growing but not with data correction or two more dimensions growing simultaneously.
% \edgar{What does "not extensively-studied" mean? To what degree do the above works adress this problem and what is the gap?}

To fill the gap, we propose \method~-- a general framework that deals with the above-mentioned online tensor updating scenarios. 
% {\color{blue} \qc{I would suggest to delete the blue part here and move it to Section 2.}
% %, which can factorize/complete such complex tensor considering various evolving patterns.   
% Specifically, given the data tensor $\mathcal{X}^{t}$ and its mask $\Omega^{t}$ (an index set of the observed entries) at time $t$, we aim at tracking the factorization of $\mathcal{X}^t$ and perform completion of its missing entries over time by assuming a low-rank structure in $\mathcal{X}^t$. Let $\mathcal{Y}^t$ be the best low-rank approximation of $\mathcal{X}^t$. The \method estimates $\mathcal{Y}^t$ through solving
% %the underlying tensor with a low-rank approximation $\mathcal{Y}^t$, such that the following loss is minimized,
% % \begin{align} 
% % \label{eq:intro_form}
% % &\mathcal{L}_{\Omega^t}(\mathcal{X}^t;\mathcal{Y}^t),~~~~s.t.~~\mathcal{Y}^t~\mbox{has low CP-rank},
% % \end{align}
% \begin{align}\label{eq:intro_form}
% \min_{\mathrm{Y}^t}~\mathcal{L}_{\Omega^t}(\mathcal{X}^t;\mathcal{Y}^t),\;\;\mathrm{s.~t.}\;\;\mathrm{rank}(\mathcal{Y}^t) \leq K,
% \end{align}
% where $\mathcal{L}_{\Omega^t}(\cdot)$ is an objective function defined on $\Omega^t$ and $\mathrm{rank}(\cdot)$ denotes the rank of its objective.
% }
The major contributions of the paper are summarized as follows:
\begin{itemize}
	\item We propose a unified framework \method for online tensor factorization/completion with complex evolving patterns such as mode growth, missing filling, and value update, while previous models cannot handle such general settings.
% \edgar{Specify the patterns and also what aspect of that (being able to handle some new pattern or a combination of patterns) is novel}
    \vspace{-5mm}
	\item We propose $\mbox{\method}_{\light}$, i.e., a more memory and computational efficient version of \method, to deal with cases where historical tensor elements are unavailable due to limited storage or privacy-related issues. 
	
% 	\item We show that both \method and $\mbox{\method}_{\light}$ are able to handle a combination of online tensor challenges efficiently. %have comparable space and time complexity as previous methods.
% 	\edgar{The first part seems repetitive with contribution 1}
    \vspace{-1mm}
	\item We experimentally show that both \method and $\mbox{\method}_{\light}$ work well under a combination of online tensor challenges. The \method improves the fitness by up to $2.8\%$ and $9.2\%$ on real-world Covid and medical claim datasets, respectively. In comparison, $\mbox{\method}_{\light}$ provides comparable fitness scores as \method with $20\%+$ complexity reduction
% 	and XXX\% memory efficiency 
	compared to the baseline methods.
% 	\qc{I wrote XXX for memory efficiency. If you don't have time to come up such a result before the deadline, please delete it.}
% 	\edgar{This experimental result summary is too general, mention what comparisons were done and try to give a numerical summary of improvement attained (achieving accuracy improvements of a-zX)}
\end{itemize}

The first version of GOCPT package has been released in PyPI\footnote{https://pypi.org/project/GOCPT/} and open-sourced in GitHub\footnote{https://github.com/ycq091044/GOCPT}. Supplementary of this paper can be found in the same GitHub repository.

%% file: sec2-problem.tex
\vspace{-2mm}
\section{Problem Formulation}
\paragraph{Notations.} 
We use plain letters for scalars, e.g., $x$ or $X$, boldface uppercase letters for matrices, e.g., $\mathbf{X}$, boldface lowercase letters for vectors, e.g., $\mathbf{x}$, and Euler script letters for tensors or sets, e.g., $\mathcal{X}$. Tensors are multidimensional arrays indexed by three or more modes. For example, an $N_{th}$-order tensor $\mathcal{X}$ is an $N$-dimensional array of size $I_1\times I_2\times\cdots\times I_N$, where $x_{i_1i_2 \cdots i_N}$ is the element at the $(i_1,i_2,\cdots,i_N)$ location. 
For a matrix $\mathbf{X}$, the $r$-th row is denoted by $\mathbf{
x}_{r}$. We use $\oast$ for Hadamard product (element-wise product), $\odot$ for Khatri-Rao product, and $\llbracket\cdot\rrbracket$ for Kruskal product  (which inputs matrices and outputs a tensor).
% For matrices $\{\mathbf{A}^n\}_{n=1}^N$, we define a series of operations as (using $\odot$ for example)
% \begin{equation*}
%     \odot_{k\neq n}\mathbf{A}^k:=\mathbf{A}^1\odot\cdots\odot \mathbf{A}^{n-1}\odot \mathbf{A}^{n+1}\odot\cdots\odot \mathbf{A}^N.
% \end{equation*} The same convention works for $\oast$ if the dimensions are valid for the corresponding operation. 

For an incomplete observation of tensor $\mathcal{X}$, we use a mask tensor $\Omega$ to indicate the observed entries: if $x_{i_1i_2\cdots i_N}$ is observed, then $\Omega_{i_1i_2\cdots i_N}=1$, otherwise $0$. Thus, $\mathcal{X}\oast\Omega$ is the actual observed data. In this paper, $\Omega$ can also be viewed as an index set of the observed entries. We define $\|(\cdot)_\Omega\|_F^2$ as the sum of element-wise squares restricted on $\Omega$, e.g.,
\begin{equation*}
	\left\|(\mathcal{X}-\mathcal{Y})_\Omega\right\|_F^2\equiv \sum_{(i_1,\cdots,i_N)\in\Omega}(x_{i_1\cdots i_N}-y_{i_1\cdots i_N})^2,
\end{equation*}
where $\mathcal{X}$ and $\mathcal{Y}$ may not necessarily be of the same size, but $\Omega$ must index within the bounds of both tensors.
% The mask $\Omega$ can be viewed as an index set or a tensor where there is no ambiguity. 
% \edgar{Change above sentence to define $\Omega$ directly and define it where it is first used rather than here.}
We describe basic matrix/tensor algebra in appendix~A, where we also list a table to summarize all the notations used in the paper.
% \qc{same question here, if X and Y have different shape, how Omega records their sample indices?}

\subsection{Problem Definition} \label{sec:problem-formulation}
% \begin{figure}[htbp!] \centering
% 	\includegraphics[width=2.5in]{figure/8_decompose.png} \caption{Eight Blocks Decomposition} \label{fig:8-decompose} 
% \end{figure}

\paragraph{Modeling Streaming Tensors.}
Real-world streaming data comes with indexing features and quantities, for example, we may receive a set of disease count tuples on a daily basis, e.g., (location, disease, date; count), where the first three features can be used to locate in a third-order tensor and the counts can be viewed as tensor elements.

\begin{figure}[htbp!] \centering
	\includegraphics[width=\linewidth]{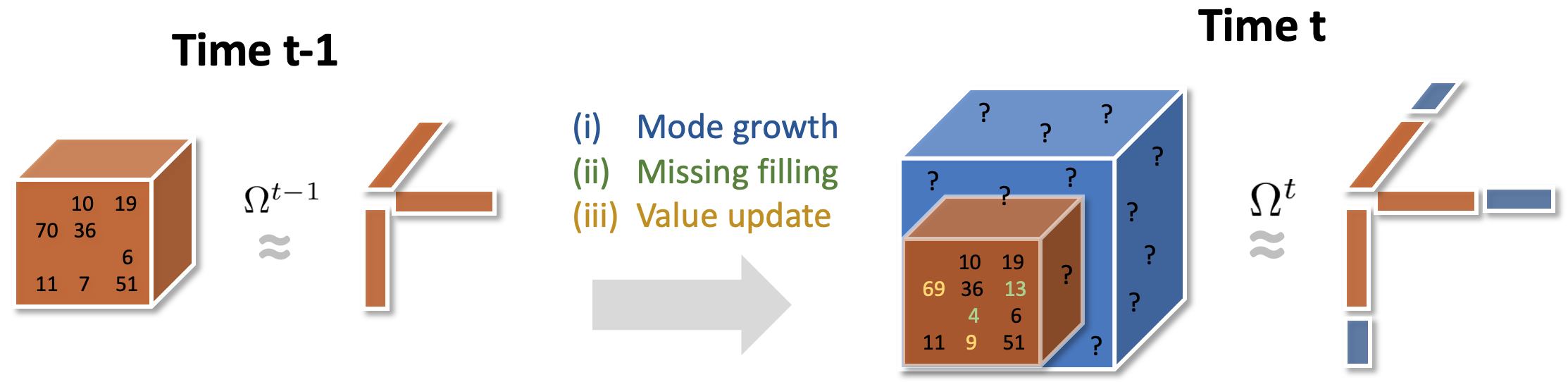} \caption{Illustration of Online Tensor Evolution} \label{fig:data-evolve} 
\end{figure}

Formally, a streaming of index-element tuples, e.g., represented by $(i_1,\cdots,i_N;~x_{i_1\cdots i_N})$, can be modeled as an evolving tensor structure. This paper considers three typical types of evolution, shown in Fig.~\ref{fig:data-evolve}. 

\begin{itemize}
	\item {\bf (i) Mode growth.} New (incomplete) slices are added along one or more modes. Refer to the blue parts. 
% 	In fact, \edgar{what formulation? our subsequent algorithm? it seems the discussion of shrinking modes is needless generality that is not evaluated}
% 	this formulation can also include the cases where mode shrinks (e.g., old dimension is not into consideration) and the size of the corresponding factor will shrink also. We model the mode growth since it is more common and mode shrink can be modeled similarly.
    \vspace{-1mm}
	\item {\bf (ii) Missing filling.} Some missing values in the old tensor is received. Refer to the green entries in the figure.
	\vspace{-1mm}
	\item {\bf (iii) Value update.} Previously observed entries may change due to new information. Refer to yellow entries. 
% 	\edgar{Not a sentence, should also make references to color consistent, and it is probably better to instead have them in the figure caption. As is, the figure may not be accessible to colorblind readers however.}
% 	\qc{I agree with Edgar.}
\end{itemize}
To track the evolution process, this paper proposes a general framework for solving the following problem on the fly.

\begin{problem}[Online Tensor Factorization and Completion] \label{problem:1}
    Suppose tensor $\mathcal{X}^{t-1}\in\mathbb{R}^{I_1^{t-1}\times\cdots\times I_N^{t-1}}$ admits a low-rank approximation $\mathcal{Y}^{t-1}$ at time $(t-1)$, parametrized by rank-$R$ factors $\{\mathbf{A}^{n,t-1}\in\mathbb{R}^{I_n^{t-1}\times R}\}_{n=1}^N$, i.e., $\mathcal{Y}^{t-1}=\llbracket\mathbf{A}^{1,t-1},\cdots,\mathbf{A}^{N,t-1}\rrbracket$. Given the mask $\Omega^{t-1}$, it satisfies
    \begin{equation} \label{eq:result-time-t-1}
        \Omega^{t-1}\oast\mathcal{X}^{t-1} \approx \Omega^{t-1}\oast\mathcal{Y}^{t-1}. 
    \end{equation}
	Following the evolution patterns, given the new underlying tensor  $\mathcal{X}^t\in\mathbb{R}^{I_1^t\times\cdots\times I_N^t}$ and the mask $\Omega^{t}$ at time $t$, our target is to find $N$ new factor matrices $\mathbf{A}^{n,t}\in\mathbb{R}^{I_n^t\times R}$, $I_n^t\geq I_n^{t-1}, ~\forall n$ and define $\mathcal{Y}^t=\llbracket\mathbf{A}^{1,t},\cdots,\mathbf{A}^{N,t}\rrbracket$, so as to minimize,
	\begin{align} \label{eq:original_generalized_objective}
		\mathcal{L}_{\Omega^t}(\mathcal{X}^t;\mathcal{Y}^t)
		\equiv&\sum_{(i_1,\cdots,i_N)\in\Omega^t} l({x}^t_{i_1\cdots i_N};{y}^t_{i_1\cdots i_N}),
	\end{align}
	where $l(\cdot)$ is an element-wise
loss function and the data and parameters are separated by semicolon.
\end{problem}

\subsection{Regularization and Approximation} \label{sec:regularization}
% \qc{Why don't we write the regularization terms in (3) and explain them after `where'? The regularization terms are common, it's unnecessary to spend a subsection to explain them. We can focus more on the approximation here.}
This section expands the objective defined in Eqn.~\eqref{eq:original_generalized_objective} with two regularization objectives.

\begin{itemize}
    \item {\bf Regularization on Individual Factors.} 
    % Common regularizations used in the literature, e.g., L2 norm \cite{yang2021augmented} and nuclear norm \cite{song2017multi}, can show various benefits in tensor related optimization problems, e.g., preventing overfitting and controling the scaling issue of factor matrices. 
    We add a generic term $\mathcal{L}_{reg}(\{\mathbf{A}^{n,t}\})$ to Eqn.~\eqref{eq:original_generalized_objective}, and it can be later customized based on application background.
    % \edgar{This discussion would be more useful if Lreg is already defined, or as direct context prior to defining Lreg. This related work discussion does not contrast our work to prior art.}
    \item {\bf Regularization on the Reconstruction.}
    Usually, from time $(t-1)$ to $t$, the historical part of the tensor will not change much, and thus we 
    % relax the proximal regularization and 
    assume that the new reconstruction $\mathcal{Y}^t$, restricted on the bound of previous tensor, will not change significantly from the previous $\mathcal{Y}^{t-1}$.
    \begin{equation*}
        \mathcal{L}_{\rec}(\mathcal{Y}^{t-1};\mathcal{Y}^t)=\sum_{1\leq i_n\leq I_n^{t-1},\forall n} l(y^{t-1}_{i_1\cdots i_N};y^{t}_{i_1\cdots i_N}).
    \end{equation*}
    %Proximal regularization \cite{do2009proximal} has been widely used to constrain the change of individual parameters, which is defined by a quadratic term over the difference of the new and old parameters. 
\end{itemize}
Considering these two regularizations, the generalized objective can be written as,
    % \begin{equation} \label{eq:generalized_objective}
    %     \begin{aligned}
    %     \mathcal{L} =& \sum_{(i_1,\cdots,i_N)\in\Omega^t} l({x}^t_{i_1\cdots i_N};{y}^t_{i_1\cdots i_N}) \\
    %     &+\alpha\sum_{1\leq i_n\leq I_N^{t-1},\forall n}l(y^{t-1}_{i_1\cdots i_N};y^{t}_{i_1\cdots i_N}) \\
    %     &+\beta\mathcal{L}_{reg}(\{\mathbf{A}^{n,t}\}),
    %      \end{aligned}
    % \end{equation}
    \begin{equation} \label{eq:generalized_objective}
        \mathcal{L} = \mathcal{L}_{\Omega^t}(\mathcal{X}^t;\mathcal{Y}^t) +\alpha\mathcal{L}_{\rec}(\mathcal{Y}^{t-1};\mathcal{Y}^t) + \beta\mathcal{L}_{reg}(\{\mathbf{A}^{n,t}\}),
    \end{equation}
    where $\alpha,\beta$ are (time-varying) hyperparameters.
    
\paragraph{Approximation.}

% \qc{The context here is part of the proposed method. If someone did it before, please add a reference, otherwise, should we move them to Section 3 as a major novelty?}

In Eqn.~\eqref{eq:original_generalized_objective}, the current tensor data masked by $\Omega^t$ consists of two parts: (i) {\bf old unchanged data} (indicating dark elements in Fig.~\ref{fig:data-evolve}), we denote it by $\Omega^{t,\old}$, which is a subset of $\Omega^{t-1}$; (ii) {\bf newly added data} (blue part, green and yellow entries in Fig.~\ref{fig:data-evolve}), denoted by set subtraction $\tilde{\Omega}^t=\Omega^t\setminus\Omega^{t,\old}$. 

The old unchanged data can be in large size. Sometime, this part of data may not be entirely preserved  due to (i)  limited memory footprint or (ii) privacy related issues. By replacing the old tensor with its reconstruction \cite{song2017multi}, we can avoid the access to the old data. Thus, we consider an approximation for the first term of Eqn.~\eqref{eq:generalized_objective},
\begin{equation} \label{eq:the-approximation} \small
    \begin{aligned}
            \mathcal{L}_{\Omega^t}(\mathcal{X}^t;\mathcal{Y}^t) =~& \mathcal{L}_{\tilde{\Omega}^t}(\mathcal{X}^t;\mathcal{Y}^t)+\mathcal{L}_{\Omega^{t,\old}}(\mathcal{X}^t;\mathcal{Y}^t)\\
            =~& \mathcal{L}_{\tilde{\Omega}^t}(\mathcal{X}^t;\mathcal{Y}^t)+\mathcal{L}_{\Omega^{t,\old}}(\mathcal{X}^{t-1};\mathcal{Y}^t) \\
            \approx~& \mathcal{L}_{\tilde{\Omega}^t}(\mathcal{X}^t;\mathcal{Y}^t) +\mathcal{L}_{\Omega^{t,\old}}(\mathcal{Y}^{t-1};\mathcal{Y}^t) \\
            =~& \mathcal{L}_{\tilde{\Omega}^t}(\mathcal{X}^t;\mathcal{Y}^t) +\sum_{\Omega^{t,\old}} l({y}^{t-1}_{i_1\cdots i_N};{y}^t_{i_1\cdots i_N}).
    \end{aligned}
\end{equation}

In the above derivation, the rationale of the approximation is the result in Eqn.~\eqref{eq:result-time-t-1}. In Eqn.~\eqref{eq:the-approximation}, we find that the term $\sum_{\Omega^{t,\old}} l({y}^{t-1}_{i_1\cdots i_N};{y}^t_{i_1\cdots i_N})$ is part of the reconstruction regularization $\mathcal{L}_{\rec}(\mathcal{Y}^{t-1};\mathcal{Y}^t)$ and thus can be absorbed. Therefore, we can use  $\mathcal{L}_{\tilde{\Omega}^t}(\mathcal{X}^t;\mathcal{Y}^t)$ to replace the full quantity $\mathcal{L}_{\Omega^t}(\mathcal{X}^t;\mathcal{Y}^t)$ in Eqn.~\eqref{eq:generalized_objective}, which results in {\bf a more efficient objective for streaming data processing},
	\begin{equation} \label{eq:light_generalized_objective} \small
	\begin{aligned}
        \mathcal{L}_{\light} = \mathcal{L}_{\tilde{\Omega}^t}(\mathcal{X}^t;\mathcal{Y}^t) +\alpha\mathcal{L}_{\rec}(\mathcal{Y}^{t-1};\mathcal{Y}^t) + \beta\mathcal{L}_{reg}(\{\mathbf{A}^{n,t}\}).
        \end{aligned}
    \end{equation}
	For this new objective, the unchanged part $\Omega^{t,\old}$ is not counted in the first term (only the new elements $\tilde{\Omega}^t$ counted), however, it is captured implicitly in the second term. 
	
	In sum, $\mathcal{L}$ and $\mathcal{L}_{\light}$ are two objectives in our framework. They have a similar expression but with different access to the tensor data (i.e., the former with mask $\Omega^t$ and the latter with $\tilde{\Omega}^t$). Generally, $\mathcal{L}$ is more accurate while $\mathcal{L}_{\light}$ is more efficient and can be applied to more challenging scenarios.

\subsection{Generalized Optimization Algorithm}
\paragraph{Structure of Parameters.} For two general objectives defined in Eqn.~\eqref{eq:generalized_objective} and \eqref{eq:light_generalized_objective}, the parameters $\{\mathbf{A}^{n,t}\in\mathbb{R}^{I_n^{t}\times R}\}$ are constructed by the upper blocks $\{\mathbf{U}^{n,t}\in\mathbb{R}^{I_n^{t-1}\times R}\}$ and the lower blocks $\{\mathbf{L}^{n,t}\in\mathbb{R}^{(I_n^t-I_n^{t-1})\times R}\}$, i.e.,
\begin{equation*} \small
	\mathbf{A}^{n,t} = \begin{bmatrix}
		\mathbf{U}^{n,t}\\
		\mathbf{L}^{n,t}
	\end{bmatrix},~\forall~n,
\end{equation*}
where $\mathbf{U}^{n,t}$ corresponds to the old dimensions along tensor mode-$n$ and $\mathbf{L}^{n,t}$ is for the newly added dimensions of mode-$n$ due to mode growth. To reduce clutter, we use ``factors" to refer to  $\mathbf{A}^{n,t}$, and ``blocks" for $\mathbf{U}^{n,t},\mathbf{L}^{n,t}$.

\paragraph{Parameter Initialization.} We use the previous factors at time $(t-1)$ to initialize the upper blocks, i.e.,
\begin{equation}
    \hat{\mathbf{U}}^{n,t} ~\stackrel{\init}{\leftarrow} ~\mathbf{A}^{n,t-1},~\forall~n. \label{eq:upper_factor-initialization}
\end{equation}

\noindent Note that, we use hat $~\hat{}~$ to denote the initialized block/factor. 

Each lower block is initialized by solving the following objective, individually, $\forall~n$,
\begin{equation} \label{eq:lower_factor-initialization} \small
    \hat{\mathbf{L}}^{n,t} ~\stackrel{\init}{\leftarrow}~ \arg \min_{\mathbf{L}^{n,t}} \sum_{\begin{subarray}{c}
    		(i_1,\cdots,i_N)\in\Omega^t\\ 1\leq i_k\leq I^{t-1}_k,~k\neq n\\
    		I^{t-1}_k < i_k \leq I^{t}_k,~k=n
    	\end{subarray}} l({x}^t_{i_1\cdots i_N};{y}^t_{i_1\cdots i_N}).
\end{equation} 
Here, $\mathcal{Y}^t$ is constructed from the parameters $\{\mathbf{A}^{n,t}\}$, and $i_k$ is the row index of the mode-$k$ factor $\mathbf{A}^{k,t}$. The summation is over the indices bounded by the {\em intersection} of $\Omega^t$ and an $N$-dim cube, where other $N-1$ modes use the old dimensions and mode-$n$ uses the new dimensions. Thus, this objective essentially uses $\{\hat{\mathbf{U}}^{k,t}\}_{k\neq n}$ to initialize $\mathbf{L}^{n,t}$.

\paragraph{Parameter Updating.} Generally, for any differentiable loss $l(\cdot)$, e.g, Frobenius norm \cite{yang2021mtc} or KL divergence \cite{hong2020generalized}, we can apply gradient based methods, to update the factor matrices. The choices of loss function, regularization term, optimization method can be customized based on the applications.

%% file: sec3-method.tex
\section{\method Framework} \label{sec:baseline_framework}
% \vspace{-1mm}
\subsection{\method Objectives} 
To instantiate a specific instance from the general algorithm formulation, we present \method with
\begin{itemize}
    \item {\em Squared residual loss}: $l(x;y) = (x-y)^2$;
    \item {\em L2 regularization}: $\mathcal{L}_{reg}(\{\mathbf{A}^{n,t}\})= \sum_{n=1}^N\left\|\mathbf{A}^{n,t}\right\|_F^2$;
    \item {\em Alternating least squares optimizer (ALS):} updating factor matrices in a sequence by fixing other factors.
\end{itemize}
Then, the general objectives in Eqn.~\eqref{eq:generalized_objective} \eqref{eq:light_generalized_objective} becomes
% \vspace{-1mm}
\begin{equation} \label{eq:baseline_objective} \small 
    \begin{aligned}
   &\mathcal{L}_{\light} =
   \left\|(\mathcal{X}^t-\llbracket\mathbf{A}^{1,t},\cdots,\mathbf{A}^{N,t}\rrbracket)_{\tilde{\Omega}^t}\right\|_F^2 \\
   &\quad+\alpha\left\|\mathcal{Y}^{t-1}-\llbracket\mathbf{U}^{1,t},\cdots,\mathbf{U}^{N,t}\rrbracket\right\|_F^2 +\beta \sum_{n=1}^N\left\|\mathbf{A}^{n,t}\right\|_F^2,
   \end{aligned}
\end{equation}
while the form of $\mathcal{L}$ is identical but with a different mask $\Omega^t$. Here, $\mathcal{L}_{\light}$ has only access to the new data $\{{x}^t_{i_1\cdots i_N}\}_{\tilde{\Omega}^t}$ but $\mathcal{L}$ has full access to the entire tensor $\{{x}^t_{i_1\cdots i_N}\}_{\Omega^t}$ up to time $t$. 

In this objective, the second term (regularization on the reconstruction defined in Sec.~\ref{sec:regularization}) is restricted on $\{(i_1,\cdots,i_N):1\leq i_n\leq I_n^{t-1},\forall n\}$, so only the upper blocks $\mathbf{U}^{n,t}\in\mathbb{R}^{I_n^{t-1}\times R}$ of factors $\mathbf{A}^{n,t}\in\mathbb{R}^{I_n^t\times R},~\forall n$, are involved. Note that, $\mathcal{L}$ and $\mathcal{L}_{\light}$ are optimized following the same procedures.

% \qc{The performance with $L_E$ is much worse than $L$ and several baselines. The only advantage seems to be the complexity. Someone may ask `does it deserve to have the complexity improvement (a few seconds) at the cost of sacrificing accuracy? For me, I didn't see the value of $L_E$.}

\subsection{\method Optimization Algorithm}\label{sec:rw-als}

For our objectives in Eqn.~\eqref{eq:baseline_objective}, we outline the optimization flow: we first {\bf initialize} the factor blocks; next, we {\bf update} the upper or lower blocks (by fixing other blocks) following this order: $\mathbf{U}^{1,t},\mathbf{L}^{1,t},\cdots,\mathbf{U}^{N,t},\mathbf{L}^{N,t}$. 

% \vspace{-0.5mm}
\paragraph{Factor Initialization.} As mentioned before, the upper blocks, $\{\mathbf{U}^{n,t}\}_{n=1}^N$, are initialized in Eqn.~\eqref{eq:upper_factor-initialization}.
In this specific framework, the minimization problem for initializing lower blocks $\{\mathbf{L}^{n,t}\}_{n=1}^N$ (in Eqn.~\eqref{eq:lower_factor-initialization}) can be reformed as
\begin{equation}\label{eq:baseline-initialize-lower} \small
   \arg\min_{\mathbf{L}^{n,t}}
   \left\|(\mathcal{X}^{t}-\llbracket\cdots,\hat{\mathbf{U}}^{n-1,t},\mathbf{L}^{n,t},\hat{\mathbf{U}}^{n+1,t},\cdots\rrbracket)_{\Omega^{n,t}}\right\|_F^2,
\end{equation}
where $\Omega^{n,t}$ denotes the {\em intersection} space in Eqn.~\eqref{eq:lower_factor-initialization}. The initialization problem in Eqn.~\eqref{eq:baseline-initialize-lower} and the targeted objectives in Eqn.~\eqref{eq:baseline_objective} can be consistently handled by the following.

% \vspace{-0.5mm}
\paragraph{Factor Updating Strategies.} Our optimization strategy depends on the density of tensor mask, e.g., $\tilde{\Omega}^t$ in Eqn.~\eqref{eq:baseline_objective}. For an evolving tensor with sparse new updates (for example, covid disease count tensor, where the whole tensor is sparse and new disease data or value correction comes irregularly at random locations), we operate only on the new elements by {\em sparse strategy}, while for tensors with dense updates (for example, multiangle imagery tensors are collected real-time and update slice-by-slice while each slice may have some missing or distortion values), we first impute the tensor to a full tensor, then apply dense operations by our {\em dense strategy}.
For example, we solve Eqn.~\eqref{eq:baseline_objective} as follows:
% \footnote{%It is hard to separate dense versus sparse by a hard threshold. 
% In appendix~\ref{sec:ablation-study}, we conduct an ablation study for scenarios with different mask density. 
% {\em In our experiments, we use the sparse strategy for the general case and the tensor completion case and use dense strategy for the factorization case}.}:
% \vspace{-0.5mm}
\begin{itemize}
    \item {\bf Sparse strategy.}
 If the $\tilde{\Omega}^t$ (the index set of newly added data at time $t$) is sparse, then we extend the CP completion alternating least squares (CPC-ALS) \cite{10.1016/j.parco.2015.10.002} and solve for each row of the factor.

\smallskip
Let us focus on $\mathbf{a}_{i_n}^{n,t}\in\mathbb{R}^{1\times R}$, which is the $i_n$-th row of factor $\mathbf{A}^{n,t}$. To calculate its derivative, we define the intermediate variables,
 \begin{align*}
     &\mathbf{P}^{n,t}_{i_n} = \sum_{(i_1,\cdots,i_{n-1},i_{n+1},\cdots,i_N)\in\tilde{\Omega}^{t,i_n}}
     \!\!\!\!\!\!\!\!\!\!\!\!(\odot_{k\neq n}\mathbf{a}_{i_k}^{k,t})^\top(\odot_{k\neq n}\mathbf{a}_{i_k}^{k,t}) \notag\\
     &\quad+ ~\alpha\left(\oast_{k\neq n}{\mathbf{U}^{k,t}}^\top\mathbf{U}^{k,t}\right) + \beta\mathbf{I},~~~\forall n,~\forall i_n,\\
     &\mathbf{q}^{n,t}_{i_n} = \sum_{(i_1,\cdots,i_{n-1},i_{n+1},\cdots,i_N)\in\tilde{\Omega}^{t,i_n}}x^t_{i_1,\cdots,i_N}(\odot_{k\neq n}\mathbf{a}_{i_k}^{k,t}) \notag\\
     &\quad+~ \alpha\mathbf{a}^{n,t-1}_{i_n}\left(\oast_{k\neq n}{\mathbf{A}^{k,t-1}}^\top\mathbf{U}^{k,t}\right),~~~\forall n,~\forall i_n.
 \end{align*}
 Here, $\tilde{\Omega}^{t,i_n}$ (we slightly abuse the notation) indicates the $i_n$-th slice of mask $\tilde{\Omega}^t$ along the $n$-th mode, and 
  \begin{equation*}
    \begin{aligned}
    \odot_{k\neq n}\mathbf{a}_{i_k}^{k,t}\equiv\mathbf{a}_{i_1}^{1,t}\odot\cdots\odot \mathbf{a}_{i_{n-1}}^{n-1,t}\odot \mathbf{a}_{i_{n+1}}^{n+1,t}\odot\cdots\odot \mathbf{a}_{i_{N}}^{N,t}.
    \end{aligned}
\end{equation*}
The same convention works for $\oast$. Here, 
$\mathbf{P}^{n,t}_{i_n}\in\mathbb{R}^{R\times R}$ is a positive definite matrix and $\mathbf{q}^{n,t}_{i_n}\in\mathbb{R}^{1\times R}$.
 
 \smallskip
 Then, the derivative w.r.t. each row can be expressed as,
 \begin{align*}
     \frac{\partial\mathcal{L}_{\light}}{\partial \mathbf{a}^{n,t}_{i_n}} =~ 2\mathbf{a}^{n,t}_{i_n}\mathbf{P}^{n,t}_{i_n} - 2\mathbf{q}^{n,t}_{i_n},~\forall n,~\forall i_n\in[1,\cdots,I_n^{t-1}],
 \end{align*}
and it applies to $\forall~i_n\in[I_n^{t-1}+1,\cdots,I_n^{t}]$ with $\alpha=0$.
 
 \smallskip
 Next, given that the objective is a quadratic function w.r.t. $\mathbf{a}^{n,t}_{i_n}$, we set the above derivative to zero and use the global minimizer to update each row,
 \begin{equation} \label{eq:sparse-solution}
     \mathbf{a}^{n,t}_{i_n}~~\stackrel{\update}{\leftarrow}~~\mathbf{q}^{n,t}_{i_n}(\mathbf{P}^{n,t}_{i_n})^{-1},~~~\forall n,~\forall i_n.
 \end{equation}
 Note that, this row-wise updating can apply in parallel. If $\tilde{\Omega}^{t,i_n}$ is empty, then we do not update $\mathbf{a}^{n,t}_{i_n}$.

\item {\bf Dense strategy.} If $\tilde{\Omega}^t$ is dense, then we extend the EM-ALS \cite{acar2011scalable} method, which applies standard ALS algorithm on the imputed tensor. To be more specific, we first impute the full tensor by interpolating/estimating the unobserved elements,
    \begin{equation*}
        \hat{\mathcal{X}}^t = \tilde{\Omega}^t\oast\mathcal{X}^t + (1-\tilde{\Omega}^t)\oast\llbracket\hat{\mathbf{A}}^{1,t},\cdots,\hat{\mathbf{A}}^{N,t}\rrbracket,
    \end{equation*}
    where we use $\hat{\mathcal{X}}^t$ to denote the estimated full tensor, and $\{\hat{\mathbf{A}}^{n,t}\}_{n=1}^N$ are the initialized factors. Then, the first term of the objective in Eqn.~\eqref{eq:baseline_objective} is approximated by a quadratic form w.r.t. each factor/block,
    \begin{equation*}
    \begin{aligned}
    \left\|\hat{\mathcal{X}}^t-\llbracket\mathbf{A}^{1,t},\cdots,\mathbf{A}^{N,t}\rrbracket\right\|_F^2.
    \end{aligned}
    \end{equation*}
    To calculate the derivative, let us define
    \begin{equation*}\scriptsize
        \begin{aligned}
        \mathbf{P}_{\mathbf{U}}^{n,t} =& \left(\oast_{k\neq n}{\mathbf{A}^{k,t}}^\top\mathbf{A}^{k,t}\right) + \alpha \left(\oast_{k\neq n}{\mathbf{U}^{k,t}}^\top\mathbf{U}^{k,t}\right)+\beta\mathbf{I}, \\
        \mathbf{Q}_{\mathbf{U}}^{n,t} =& (\hat{\mathbf{X}}_n^t)_{\mathbf{U}}\left(\odot_{k\neq n}\mathbf{A}^{k,t}\right) + \alpha\mathbf{A}^{n,t-1}\left(\oast_{k\neq n}{\mathbf{A}^{k,t-1}}^\top\mathbf{U}^{k,t}\right), \\
        \mathbf{P}_{\mathbf{L}}^{n,t} =& \left(\oast_{k\neq n}{\mathbf{A}^{k,t}}^\top\mathbf{A}^{k,t}\right) +\beta\mathbf{I}, \\
        \mathbf{Q}_{\mathbf{L}}^{n,t} =& (\hat{\mathbf{X}}_n^t)_{\mathbf{L}}\left(\odot_{k\neq n}\mathbf{A}^{k,t}\right),
        \end{aligned}
    \end{equation*}
    where $\hat{\mathbf{X}}_n^t$ is the mode-$n$ unfolding of $\hat{\mathcal{X}}^t$, and $(\hat{\mathbf{X}}_n^t)_{\mathbf{U}}\in\mathbb{R}^{I_n^{t-1}\times \prod_{n\neq k}I_n^{k,t}}$ is the first $I_n^{t-1}$ rows of $\hat{\mathbf{X}}_n^t$, while $(\hat{\mathbf{X}}_n^t)_{\mathbf{L}}\in\mathbb{R}^{(I_n^{t}-I_n^{t-1})\times \prod_{n\neq k}I_n^{k,t}}$ is the the remaining $(I_n^{t}-I_n^{t-1})$ rows of $\hat{\mathbf{X}}_n^t$. Here, $\mathbf{P}_{\mathbf{U}}^{n,t},\mathbf{P}_{\mathbf{L}}^{n,t}\in\mathbb{R}^{R\times R}$ are positive definite, $\mathbf{Q}_{\mathbf{U}}^{n,t}\in\mathbb{R}^{I_n^{t-1}\times R}$ and $\mathbf{Q}_{\mathbf{L}}^{n,t}\in\mathbb{R}^{(I_n^t-I_n^{t-1})\times R}$.
    
    \smallskip
    We express the derivative of the upper and lower blocks by the intermediate variables defined above,
    \begin{equation*}
        \begin{aligned}
        \frac{\partial\mathcal{L}_{\light}}{\partial \mathbf{U}^{n,t}} &=~ 2\mathbf{U}^{n,t}\mathbf{P}^{n,t}_{\mathbf{U}} - 2\mathbf{Q}^{n,t}_{\mathbf{U}},~~~\forall n, \\
        \frac{\partial\mathcal{L}_{\light}}{\partial \mathbf{L}^{n,t}} &=~ 2\mathbf{L}^{n,t}\mathbf{P}^{n,t}_{\mathbf{L}} - 2\mathbf{Q}^{n,t}_{\mathbf{L}},~~~\forall n.
    \end{aligned}
    \end{equation*}
    
    Here, the overall objective $\mathcal{L}_{\light}$ is a quadratic function w.r.t. $\mathbf{U}^{n,t}$ or $\mathbf{L}^{n,t}$. By setting the derivative to zero, we can obtain the global minimizer for updating,
    \begin{equation} \label{eq:dense-solution}
        \begin{aligned}
        \mathbf{U}^{n,t}~~&\stackrel{\update}{\leftarrow}~~\mathbf{Q}^{n,t}_{\mathbf{U}}(\mathbf{P}^{n,t}_{\mathbf{U}})^{-1},~~~\forall n, \\
        \mathbf{L}^{n,t}~~&\stackrel{\update}{\leftarrow}~~\mathbf{Q}^{n,t}_{\mathbf{L}}(\mathbf{P}^{n,t}_{\mathbf{L}})^{-1}, ~~~\forall n.
        \end{aligned}
    \end{equation}
\end{itemize}

% \paragraph{Block-wise Objective.} In fact, the derivative calculation can be conducted block-wise. Each factor has upper and lower blocks and $N$ factors can binarily identify $2^N$ tensor blocks (viewing $\mathcal{X}^t$ as the combination of $2^N$ blocks) at time $t$. Then, the overall objective can be viewed as a summation of $2^N$ smaller objectives with $2\times N$ shared factors (i.e, $\{\mathbf{U}^{n,t}\}_{n=1}^N, \{\mathbf{L}^{n,t}\}_{n=1}^N$), where each of the objective block can have their own decisions on either using Eqn.~\eqref{eq:sparse-solution} or \eqref{eq:dense-solution}, which makes the derivation calculation more precise.
To sum up, the optimization flow for $\mathcal{L}_{\light}$ is summarized in Algorithm~\ref{algo:alternating}. For optimizing $\mathcal{L}$, we simply modify the algorithm by replacing the input $\{{x}^t_{i_1\cdots i_N}\}_{\tilde{\Omega}^t}$ with $\{{x}^t_{i_1\cdots i_N}\}_{\Omega^t}$.
We analyze the complexity of our \method in appendix~B.
\begin{algorithm}[!ht]\footnotesize
	\SetAlgoLined
	\textbf{Input:} $\{\mathbf{A}^{n,t-1}\}_{n=1}^N,\{{x}^t_{i_1\cdots i_N}\}_{\tilde{\Omega}^t},\alpha,\beta$;\;
	
	\textbf{Parameters:} $\{\mathbf{U}^{n,t}\}_{n=1}^N,~ \{\mathbf{L}^{n,t}\}_{n=1}^N$;\;
	
	Initialize parameters by Eqn.~\eqref{eq:upper_factor-initialization} and \eqref{eq:baseline-initialize-lower};\;
	
	\For{$n\in[1,\cdots,N]$}{
	    update $\mathbf{U}^{n,t}$ using Eqn. \eqref{eq:sparse-solution} or \eqref{eq:dense-solution};\;
	    
		update $\mathbf{L}^{n,t}$ using Eqn. \eqref{eq:sparse-solution} or \eqref{eq:dense-solution};\;}
	\textbf{Output:} new factors $\left\{\mathbf{A}^{n,t}=\begin{bmatrix}
		\mathbf{U}^{n,t} \\
		\mathbf{L}^{n,t}
	\end{bmatrix}\right\}_{n=1}^N$.
	\caption{Factor Updates for $\mathcal{L}_{\light}$ at time $t$}
	\label{algo:alternating}
\end{algorithm}

%% file: sec4-unifying.tex
\section{Unifying Previous Online Settings}
Several popular online tensor factorization/completion special cases can be unified in our framework. Among them, we focus on the {\em online tensor factorization} and {\em online tensor completion}. We show that those objectives in literature can be achieved by \method. Our experiments confirm \method can obtain comparable or better performance over previous methods in those special cases. 

\subsection{Online Tensor Factorization} \label{sec:online-tensor-factorization}
\begin{figure}[htbp!] \centering
	\includegraphics[width=2.2in]{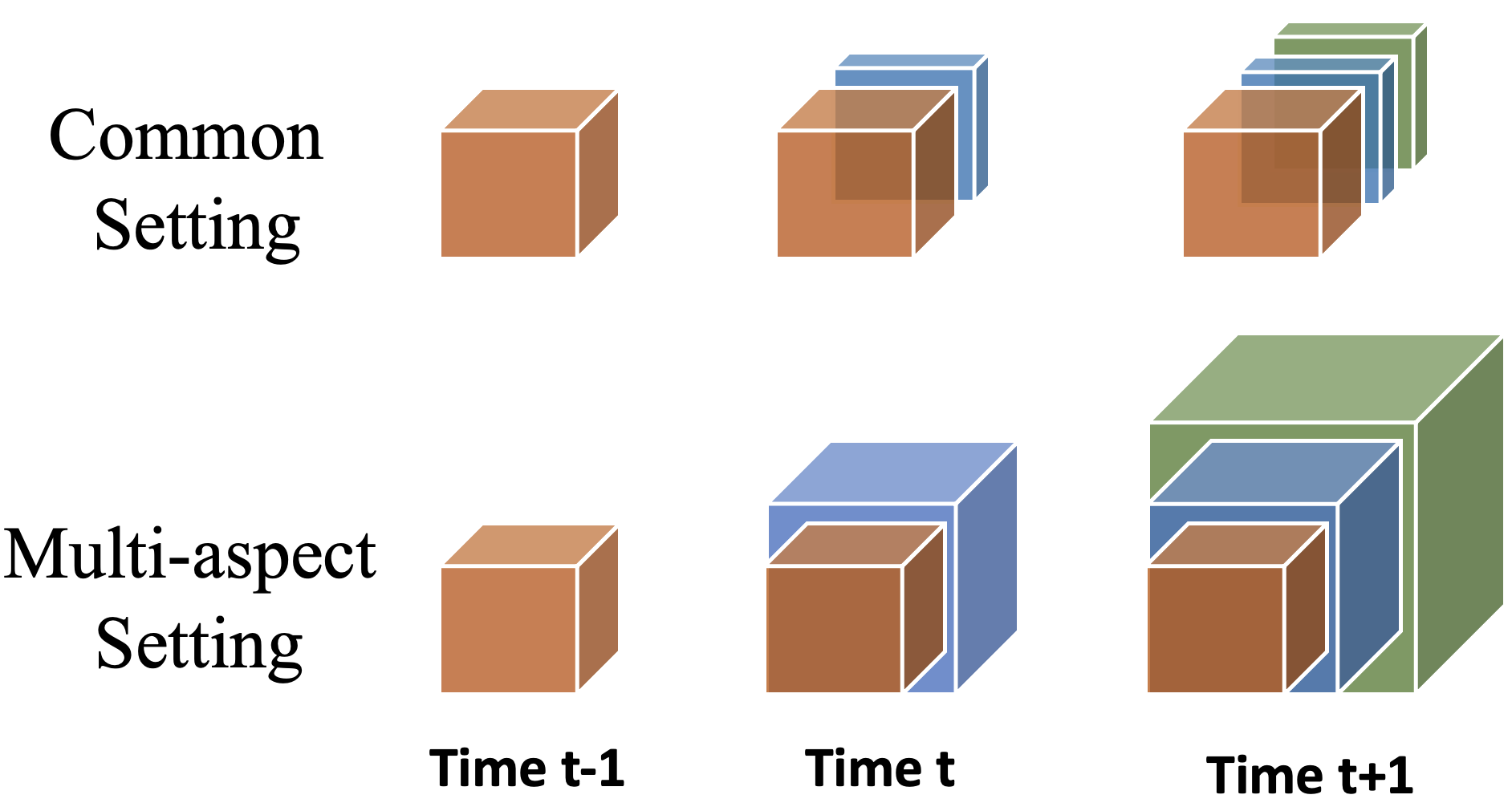} \caption{Online Tensor Factorization} \label{fig:multi-aspect} 
	\vspace{-2mm}
\end{figure}

Online tensor factorization tries to maintain the factorization results (e.g., CP factors) while the tensor is growing, shown in Fig.~\ref{fig:multi-aspect}. Zhou et al. \cite{zhou2016accelerating} proposed an accelerated algorithm for the common setting where only the temporal mode grows. \cite{song2017multi} considered the multi-aspect setting where multiple mode grows simultaneously. We discuss the multi-aspect setting below.

\begin{problem} [Multi-aspect Online Tensor Factorization]
	Suppose tensor $\mathcal{X}^{t-1}\in\mathbb{R}^{I^{t-1}_1\times I^{t-1}_2\times I^{t-1}_3}$ at time $(t-1)$ admits a low-rank approximation, induced by $\mathbf{U}^{t-1}\in\mathbb{R}^{I_1^{t-1}\times R}$,  $\mathbf{V}^{t-1}\in\mathbb{R}^{I_2^{t-1}\times R}$,
	$\mathbf{W}^{t-1}\in\mathbb{R}^{I_3^{t-1}\times R}$, such that
	\begin{equation*}
		\mathcal{X}^{t-1} \approx \llbracket\mathbf{U}^{t-1},\mathbf{V}^{t-1},\mathbf{W}^{t-1}\rrbracket.
	\end{equation*}
	
	At time $t$, we want to learn a new set of factors, $\mathbf{U}^t
	\in\mathbb{R}^{I_1^t\times R}, \mathbf{V}^t
	\in\mathbb{R}^{I_2^t\times R}, \mathbf{W}^t
	\in\mathbb{R}^{I_3^t\times R}$ to approximate the growing new tensor $\mathcal{X}^t\in\mathbb{R}^{I^{t}_1\times I^{t}_2\times I^{t}_3}$, which satisfies,
	\begin{align*}
		x_{i_1i_2i_3}^t = x_{i_1i_2i_3}^{t-1},~\forall i_n\in[1,\cdots,I_n^{t-1}],~\forall n\in[1,2,3].
	\end{align*}
\end{problem}

\paragraph{Unification.} %Both settings can be handled in our framework. For the multi-aspect setting, 
To achieve the existing objective in the literature, we simply reduce our $\mathcal{L}_{\light}$ by changing the L2 regularization into nuclear norm (i.e., the sum of singular values),
	\begin{equation*}
		\mathcal{L}_{\reg} = \gamma_1\left\|\mathbf{U}^t\right\|_* + \gamma_2\left\|\mathbf{V}^t\right\|_* + 
		\gamma_3\left\|\mathbf{W}^t\right\|_*,
	\end{equation*}
	where $\gamma_n,~\forall~n\in[1,2,3]$ are the hyperparameters and they sum up to one. This regularization is the convex relaxation of CP rank \cite{song2017multi}. Then, the objective becomes
\begin{align}
	\mathcal{L}_{\light} = & \left\|(\mathcal{X}^{t}-\llbracket\mathbf{U}^t,\mathbf{V}^t,\mathbf{W}^t\rrbracket)_{\tilde{\Omega}^t}\right\|_F^2 \notag\\
	&+\alpha\left\|(\mathcal{Y}^{t-1}-\llbracket\mathbf{U}^t,\mathbf{V}^t,\mathbf{W}^t\rrbracket)_{\Omega^{t-1}}\right\|_F^2 \notag\\
	&+\beta \left(\gamma_1\left\|\mathbf{U}^t\right\|_* + \gamma_2\left\|\mathbf{V}^t\right\|_* + 
	\gamma_3\left\|\mathbf{W}^t\right\|_*\right) \notag,
	% \label{eq:online_tensor_completion_objective},
\end{align}
where $\Omega^t$ and $\Omega^{t-1}$ are the bounds of the new and previous tensors and $\tilde{\Omega}^t=\Omega^t\setminus\Omega^{t-1}$ (in this case, $\Omega^{t,\old}=\Omega^{t-1}$) indicates the newly added elements at time $t$.
This form is identical to the objective proposed in \cite{song2017multi}. 

In experiment Sec.~\ref{sec:exp-online-tensor-factorization}, we evaluate on the temporal mode growth setting. Our methods use the L2 regularization as listed in Sec.~\ref{sec:baseline_framework} and follow Algorithm~\ref{algo:alternating}.

\subsection{Online Tensor Completion}

\begin{figure}[htbp!] \centering
	\includegraphics[width=2.0in]{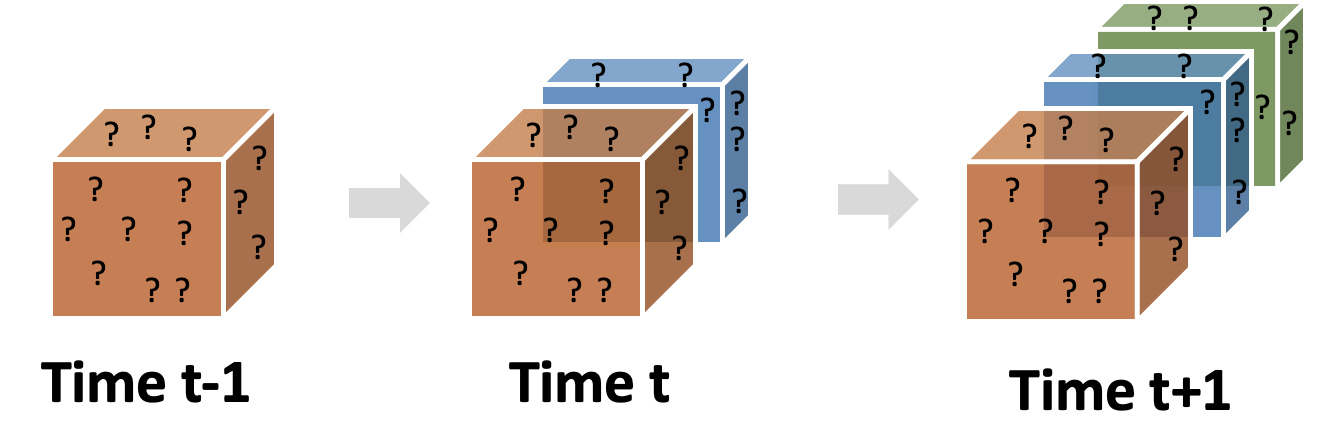} \caption{Online Tensor Completion} \label{fig:streaming-tensor-completion} 
	\vspace{-2mm}
\end{figure}

\noindent Online tensor completion studies incomplete tensors, where new incomplete slices will add to the temporal mode. The goal is to effectively compute a new tensor completion result (i.e., CP factors) for the augmented tensor. This problem is studied in \cite{mardani2015subspace,kasai2016online,kasai2019fast}. We show the formulation below.

\begin{problem}[Online Tensor Completion]
	Suppose $\mathcal{X}^{t-1}$ is the underlying tensor at time $(t-1)$, which admits a low-rank approximation, $\mathbf{U}^{t-1}\in\mathbb{R}^{I_1\times R}$,  $\mathbf{V}^{t-1}\in\mathbb{R}^{I_2\times R}$,
	$\mathbf{W}^{t-1}\in\mathbb{R}^{I^{t-1}_3\times R}$, such that given the mask $\Omega^{t-1}$,
	\begin{equation*}
		\Omega^{t-1}\oast\mathcal{X}^{t-1} \approx \Omega^{t-1}\oast\llbracket\mathbf{U}^{t-1},\mathbf{V}^{t-1},\mathbf{W}^{t-1}\rrbracket.
	\end{equation*}
	At the time $t$, given a new slice with incomplete entries, i.e., $\Omega_\Delta^t\oast\mathbf{X}_\Delta^{t}\in\mathbb{R}^{I_1\times I_2}$, the new data is concatenated along the temporal mode (the third-mode) by $\Omega^t = [\Omega^{t-1};\Omega^t_{\Delta}]$ and  $\mathcal{X}^t = [\mathcal{X}^{t-1};\mathbf{X}_\Delta^t]$.
% 	\begin{align*}
% 		\Omega^t &= [\Omega^{t-1};\Omega^t_{\Delta}]\in\mathbb{R}^{I_1\times I_2\times I_3^t},~~\Omega_{\Delta}^t\in\mathbb{R}^{I_1\times I_2}, \\
% 		\mathcal{X}^t &= [\mathcal{X}^{t-1};\mathbf{X}_\Delta^t]\in\mathbb{R}^{I_1\times I_2\times I_3^t},~~~\mathbf{X}^t_\Delta\in\mathbb{R}^{I_1\times I_2}.
% 	\end{align*}
	We want to update the approximation with $\mathbf{U}^t\in\mathbb{R}^{I_1\times R}$,  $\mathbf{V}^t\in\mathbb{R}^{I_2\times R}$, $\mathbf{W}^t\in\mathbb{R}^{I_3^t\times R}$, where $I^t_3 = I_3^{t-1}+1$, such that
	\begin{equation*}
		\Omega^t\oast\mathcal{X}^t \approx \Omega^t\oast\llbracket\mathbf{U}^t,\mathbf{V}^t,\mathbf{W}^t\rrbracket.
	\end{equation*}
\end{problem}

\paragraph{Unification.} To handle this setting, we remove the reconstruction regularizer (second term) from our $\mathcal{L}$ and 
the reduced version can be transformed into,
\begin{equation} \label{eq:streaming-tensor-completion-pre-objective} \small
    \begin{aligned}
    \mathcal{L} =&\sum_{k=1}^{I_3^{t}}\left[\alpha\left\|\left(\mathbf{X}^{t,k}-\mathbf{U}^t\diag(\mathbf{w}^t_{k})
	\mathbf{V}^{t\top}\right)_{\Omega^{t,k}}\right\|_F^2 +\beta\|\mathbf{w}_k^t\|_F^2\right]\\ &+\beta\left(\|\mathbf{U}^t\|_F^2+\|\mathbf{V}^t\|_F^2\right),
    \end{aligned}
\end{equation}
where $\Omega^{t,k}$ and $\mathbf{X}^{t,k}$ indicate the $k$-th slices of the tensor along the temporal mode, and $\mathbf{w}^t_k$ is the $k$-th row of $\mathbf{W}^t$.
From Eqn.~\eqref{eq:streaming-tensor-completion-pre-objective}, we could 
make $\alpha$ exponential decaying w.r.t. time steps (i.e., $\alpha_k=\gamma^{I_3^t-k}$), the $\beta$ within summation be time-variant (i.e., $\beta_k=\lambda_k\times \gamma^{I_3^t-k}$) and the $\beta$ outside be $\lambda_{I_3^t}$, where $\gamma,\{\lambda_k\}$ are hyperparameters. We then obtain the identical objective as proposed in \cite{mardani2015subspace}. The experiments are shown in Sec.~\ref{sec:exp-stream-tensor-completion}.

% \subsection{Unifying Other Online Settings}
% Our framework can also be used to track the CP factors in other settings, such as (i) for an incomplete tensor, its size does not change and the missing entries are gradually filled; (ii) for a complete tensor with fixed size, the tensor elements can change at each time steps. We apply our framework to these two scenarios in  appendix~\ref{app:online-tensor-completion} and \ref{exp:tensor-decomposition-changing-values}.

%% file: sec5-experiment.tex
\section{Experiments}
In the experiment, we evaluate our model on various settings.
Our models are named \method (with the objective $\mathcal{L}$) and $\mbox{\method}_{\light}$ (with objective $\mathcal{L}_{\light}$).
% We use the objective $\mathcal{L}$ (defined in Eqn.~\eqref{eq:baseline_objective}) for Sec.~\ref{sec:exp-general-online-tensor-tracking} and $\mathcal{L}_{\light}$ (defined in Eqn.~\eqref{eq:light_baseline_objective}) for Sec.~\ref{sec:exp-online-tensor-factorization} and Sec.~\ref{sec:exp-stream-tensor-completion}.
Dataset statistics are listed in Table~\ref{tb:dataset}. More details can be found in appendix~C.

\begin{table}[tbp!] \centering
	\resizebox{.42\textwidth}{!}{\centering\begin{tabular}{cccc} \toprule \textbf{Dataset} & \textbf{Format} & \textbf{Setting} \\ \midrule
	        JHU Covid & $51 \times 3\times 8 \times 209$ & General (Sec.~\ref{sec:exp-general-online-tensor-tracking})  \\
			Patient Claim & $56\times 22\times 10 \times 104$ &  General (Sec.~\ref{sec:exp-general-online-tensor-tracking})  \\
			FACE-3D & $112\times 99\times 400$ & Factorization (Sec.~\ref{sec:exp-online-tensor-factorization})  \\
			GCSS & $50\times 422\times 362$ & Factorization (Sec.~\ref{sec:exp-online-tensor-factorization})  \\
			Indian Pines & $145\times 145\times 200$ & Completion (Sec.~\ref{sec:exp-stream-tensor-completion}) \\
			CovidHT & $420 \times189\times128$ & Completion (Sec.~\ref{sec:exp-stream-tensor-completion})  \\ 
			\bottomrule \end{tabular}}
			%{\small *Prep. indicates the percentage of preparation data, which is used to obtain the initial factors at time $0$.}
	\vspace{-2.5mm}
	\caption{Data Statistics}
	\vspace{-3mm}
	%\qc{AAAI requires captions under tables or figures.}
	\label{tb:dataset} \end{table}

\subsection{Experimental Setups}
%We employ state-of-the-art {\em OnlineCPD} \cite{zhou2016accelerating}, which can only update factors once at each time step; {\em MAST} \cite{song2017multi}, which uses ADMM and ideally requires multiple updates; EM-ALS \cite{acar2011scalable,walczak2001dealing}, which first imputes the tensor and then run ALS algorithm at each time; OnlineSGD \cite{mardani2015subspace}, which computes the gradient update on the flay; OLSTEC \cite{kasai2016online,kasai2019fast}, which uses recursive least squres method for tensor tracking.

\paragraph{Metrics.} The main metric is {\bf percentage of fitness (PoF)} \cite{acar2011scalable}, which is defined for the factorization or completion problem respectively by (the higher, the better)
\begin{equation*} \scriptsize
\begin{aligned}
 1-\frac{\|\mathcal{X}-\llbracket\mathbf{A}_1\dots\mathbf{A}_N\rrbracket\|_F}{\|\mathcal{X}\|_F}~~~\mbox{or}~~~ 1-\frac{\|(1-\Omega)\oast(\mathcal{X}-\llbracket\mathbf{A}_1\dots\mathbf{A}_N\rrbracket)\|_F}{\|(1-\Omega)\oast\mathcal{X}\|_F},
\end{aligned}
\end{equation*}
where $\Omega$ and $\mathcal{X}$ are the mask and underlying tensor, $\{\mathbf{A}_1,\cdots,\mathbf{A}_N\}$ are the low-rank CP factors. We also report the {\bf total time consumption} as an efficiency indicator.

\paragraph{Baselines.} We simulate three different practical scenarios for performance comparison. Since not all existing methods can deal with the three cases, we select representative state-of-the-art algorithms in each scenario for comparison: %The baselines are different for each section.
\begin{itemize}
	\item For the {\bf general case} in Sec.~\ref{sec:exp-general-online-tensor-tracking}, most previous models cannot support this setting.
% 	\footnote{For OLSTEC and OnlineSGD, they are only designed for new incoming slices along the temporal mode, while their pipeline cannot be easily adjusted for value update and multiple mode growth. For other baselines (e.g., OnlineCPD), they work with complete tensors only without missing values.}. 
We adopt {\em EM-ALS} \cite{acar2011scalable,walczak2001dealing} and {\em CPC-ALS} \cite{10.1016/j.parco.2015.10.002} as the compared methods, which follow the similar initialization procedure in Sec.~\ref{sec:rw-als}. 
% 	We also implement the original CPC-ALS, called {\em CPC-ALS-REF}, which optimizes from scratch (i.e., randomized initialization) until convergence (max iterations exceed 50 or the percentage reconstruction difference between two consecutive iterations is less than $10^{-5}$) at time $t$. It works as a reference model and provides decent fitness at the expense of the efficiency. 
    \vspace{-1mm}
	\item For the {\bf online tensor factorization} in Sec.~\ref{sec:exp-online-tensor-factorization}, we employ {\em OnlineCPD} \cite{zhou2016accelerating}; {\em MAST} \cite{song2017multi} and {\em CPStream} \cite{smith2018streaming}, which use ADMM and require multiple iterations; {\em RLST} \cite{nion2009adaptive}, which is designed only for third-order tensors. 
	\vspace{-1mm}
	\item For the {\bf online tensor completion}  in Sec.~\ref{sec:exp-stream-tensor-completion}, we implement {\em EM-ALS} and its variant, called {\em EM-ALS (decay)}, which assigns exponential decaying weights for historical slices; {\em OnlineSGD} \cite{mardani2015subspace}; {\em OLSTEC} \cite{kasai2016online,kasai2019fast} for comparison.
\end{itemize}

 We compare the space and time complexity of each model in appendix~B. All experiments are conduct with 5 random seeds. The mean and standard deviations are reported.
 
%  by {\em Python 3.8.5}, {\em Numpy 1.19.1} and the experiments are conducted on a Linux workstation with 256 GB memory, 32 core CPUs (3.70 GHz, 128 MB cache) 

% \vspace{-1mm}
\subsection{General Case with Three Evolving Patterns} \label{sec:exp-general-online-tensor-tracking}

\paragraph{Datasets and Settings.} We use (i) JHU Covid data \cite{dong2020interactive} and (ii) proprietary Patient Claim data to conduct the evaluation. The JHU Covid data was collected from Apr. 6, 2020, to Oct. 31, 2020 and the Patient Claim dataset collected weekly data from 2018 to 2019.
% , containing 3 feature categories (new cases, deaths, and hospitalization) for 51 US states. It has 209 generation dates (GD) and 8 loading dates (LD), corresponding to tensor evolving type (i) and (ii). The Patient Claim dataset spans from 2018 to 2019 weekly, containing 22 ICD-10 disease feature categories over 56 US states, with 104 GD's and 10 LD's. 
To mimic tensor value updates on two datasets, we later add random perturbation to randomly selected $2\%$ existing data with value changes uniformly of $[-5\%,5\%]$ at each time step. 
%Specifically, suppose $x$ is one of the selected entry, we update it by $x\leftarrow x(1+s)$, where $s$ is sampled from uniform $[-5e^{-2},5e^{-2}]$ distribution. 
The leading $50\%$ slices are used as preparation data to obtain the initial factors with rank $R=5$. The results are in Table~\ref{tb:patient-claim-data}.

\begin{table}[tbp!] 
	\resizebox{0.47\textwidth}{!}{\begin{tabular}{ccccc} \toprule 
			\multirow{2}{*}{\bf Model} & \multicolumn{2}{c}{\bf JHU Covid Data} & \multicolumn{2}{c}{\bf Perturbed JHU Covid Data} \\
			\cmidrule{2-5}
			& Total Time (s) & Avg. PoF & Total Time (s) & Avg. PoF \\
			\midrule
% 			CPC-ALS-REF & 18.21 $\pm$ 0.760 & 0.7031 $\pm$ 5.493e-3 & 18.45 $\pm$ 0.510 & 0.6898 $\pm$ 6.771e-3\\
% 			\midrule
			EM-ALS & 1.68 $\pm$ 0.001 & 0.6805 $\pm$ 0.024 & 2.13 $\pm$ 0.049 & 0.6622 $\pm$ 0.047\\
        CPC-ALS & 2.14 $\pm$ 0.002 & 0.6813 $\pm$ 0.028 & 2.50 $\pm$ 0.013 & 0.6634 $\pm$ 0.021\\
        $\mbox{\method}_{\light}$ & 1.32 $\pm$ 0.004 & {\bf 0.6897 $\pm$ 0.016} & 1.72 $\pm$ 0.034 & {\bf 0.6694 $\pm$ 0.045}\\
        \method & 2.68 $\pm$ 0.002 & {\bf 0.6920 $\pm$ 0.022} & 3.17 $\pm$ 0.041 & {\bf 0.6827 $\pm$ 0.024}\\
        \bottomrule
        \toprule 
			\multirow{2}{*}{\bf Model} & \multicolumn{2}{c}{\bf Patient Claim} & \multicolumn{2}{c}{\bf Perturbed Patient Claim} \\
			\cmidrule{2-5}
			& Total Time (s) & Avg. PoF & Total Time (s) & Avg. PoF \\
			\midrule
% 			CPC-ALS-REF & 129.35 $\pm$ 3.400 & 0.5687 $\pm$ 3.060e-3 & 132.19 $\pm$ 2.200 & 0.5712 $\pm$ 4.150e-3\\
% 			\midrule
			EM-ALS & 4.37 $\pm$ 0.056 & 0.4458 $\pm$ 0.023 & 5.35 $\pm$ 0.066 & 0.4626 $\pm$ 0.021\\
			CPC-ALS & 4.74 $\pm$ 0.036 & 0.5022 $\pm$ 0.021 & 5.58 $\pm$ 0.009 & 0.5169 $\pm$ 0.019\\
			$\mbox{\method}_{\light}$ & 2.71 $\pm$ 0.033 & {\bf 0.5299 $\pm$ 0.019} & 3.27 $\pm$ 0.024 & {\bf 0.5454 $\pm$ 0.017}\\
			\method & 5.10 $\pm$ 0.037 & {\bf 0.5485 $\pm$ 0.022} & 5.91 $\pm$ 0.042 & {\bf 0.5577 $\pm$ 0.021}\\
			\bottomrule
	\end{tabular}}
	\vspace{-2mm}
	\caption{Results on General Case}\label{tb:patient-claim-data}
	%{\small* With perturbation, the performance of most models improve a little. The reason might be that the random perturbations fill some initially empty entries and may help improve the optimization.}
	\vspace{-3.7mm}
\end{table}

%\begin{figure}[ht]
%	\begin{minipage}[b]{0.47\linewidth}
%		\centering
%		\includegraphics[width=\textwidth]{figure/patient.pdf}
%	\end{minipage}
%	\begin{minipage}[b]{0.47\linewidth}
%		\centering
%		\includegraphics[width=\textwidth]{figure/patient_perturb.pdf}
%	\end{minipage}
%	\vspace{-2mm}
%	\caption{Results on (Perturbed) Patient Claim Data}
%	\label{fig:result-delayed1}
%\end{figure}

% \vspace{-1mm}
\paragraph{Result Analysis.} Overall, our models show the top fitness performance compared to the baselines, and the variant $\method_{\light}$ shows $20\%$ efficiency improvement over the best model with comparable fitness on both datasets. CPC-ALS and EM-ALS performs similarly on Covid tensor and CPC-ALS works better on the Patient Claim data, while they are inferior to our models in both fitness and efficiency.

\vspace{-1mm}
\subsection{Special Case: Online Tensor Factorization} \label{sec:exp-online-tensor-factorization}
%\paragraph{Baseline Methods.} We employ {\em OnlineCPD} \cite{zhou2016accelerating}; {\em MAST} \cite{song2017multi} and CPStream \cite{smith2018streaming}, and these two methods use ADMM and requires multiple iterations; {\em RLST} \cite{nion2009adaptive}, which is designed only for third-order tensors.
\vspace{-0.5mm}
\paragraph{Dataset and Setups.} We  present the evaluation on low-rank synthetic data. In particular, we generate three CP factors from uniform $[0,1]$ distribution and then construct a low-rank tensor $(I_1,I_2,I_3,R) = (50,50,500,5)$. We use the leading $10\%$ slices along the (third) temporal mode as preparation; then, we add one slice at each time step to simulate mode growth. We report the mean values in Figure~\ref{fig:synthetic}. 

Also, we show that our \method can provide better fitness than all baselines and $\mbox{\method}_{\light}$ outputs comparable fitness with state of the art efficiency on two real-world datasets: (i) ORL Database of Faces (FACE-3D) and (ii) Google Covid Search Symptom data (GCSS). The result tables are moved to appendix~C.3 due to space limitation.

% \vspace{-1mm}
\subsection{Special Case: Online Tensor Completion}  \label{sec:exp-stream-tensor-completion}
\begin{figure}[tbp!]
	\begin{minipage}[b]{0.47\linewidth}
		\centering
		\includegraphics[width=\textwidth]{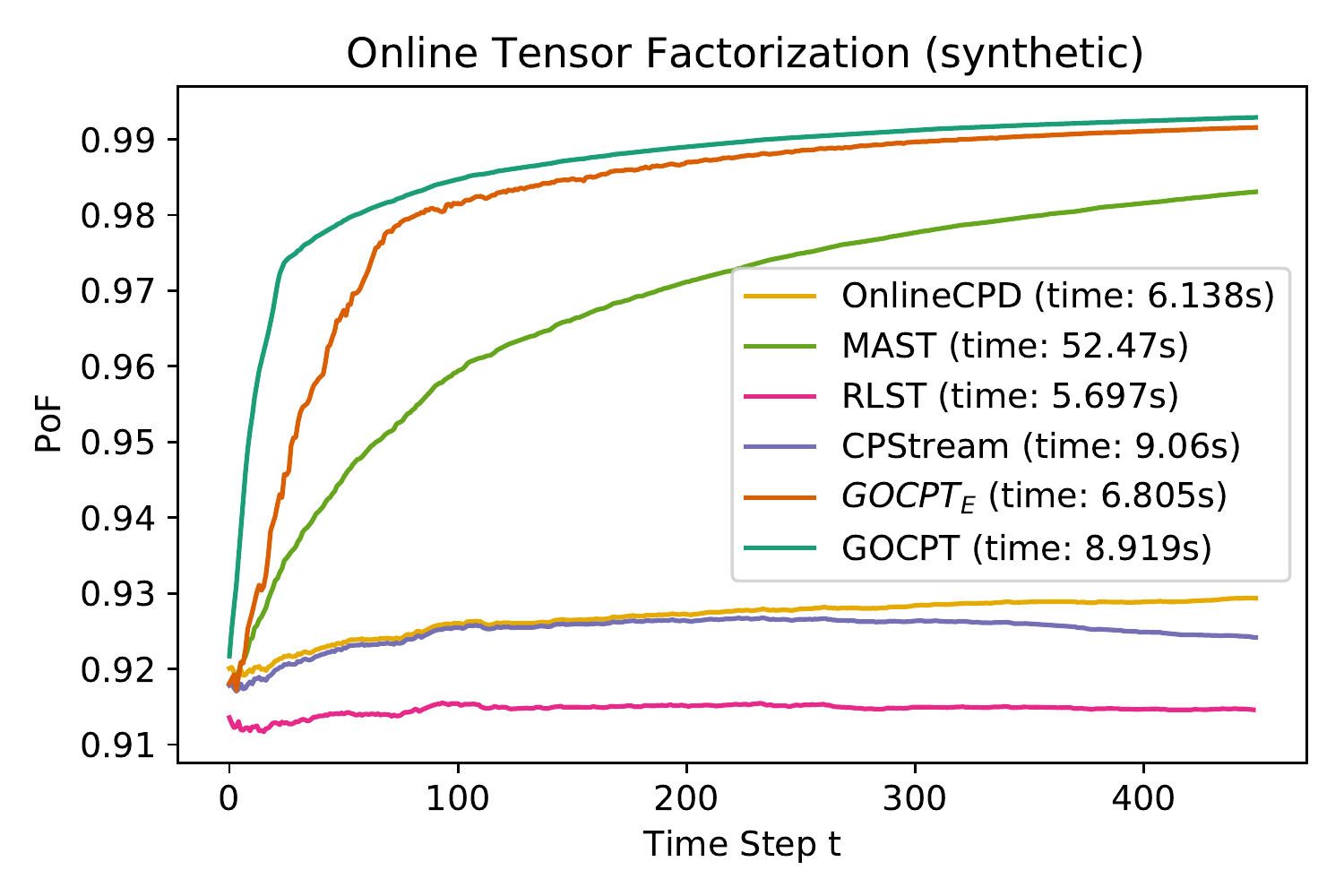}
	\end{minipage}
	\begin{minipage}[b]{0.49\linewidth}
		\centering		\includegraphics[width=\textwidth]{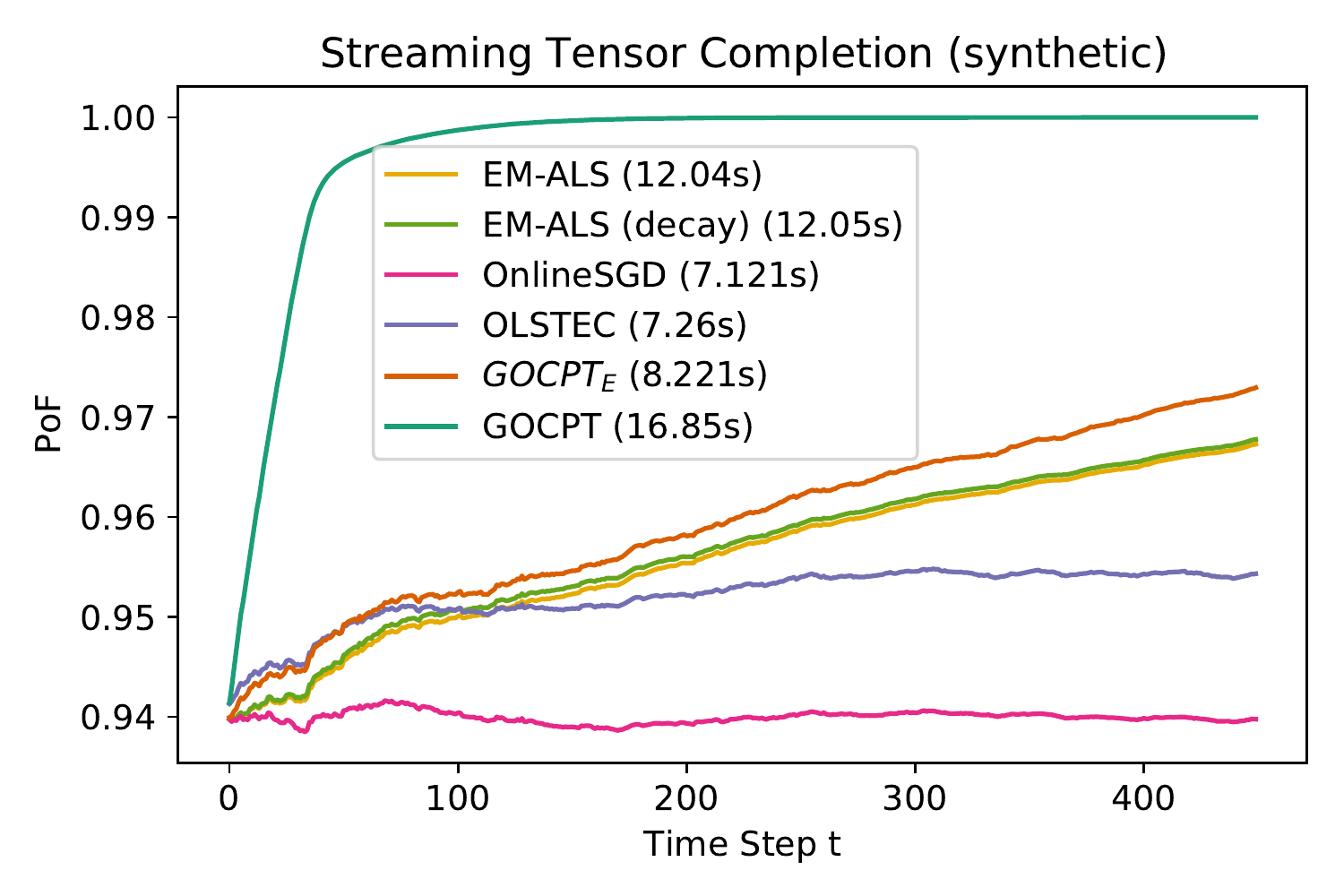}
	\end{minipage}
	\vspace{-2mm}
	\caption{Performance Comparison on Special Cases}
	\vspace{-3mm}
	\label{fig:synthetic}
\end{figure}

%\paragraph{Baseline Methods.} We consider EM-ALS \cite{acar2011scalable} and its variant, called EM-ALS (decay), which assigns exponential decaying weights for historical slices; OnlineSGD \cite{mardani2015subspace}; OLSTEC \cite{kasai2016online,kasai2019fast} as the comparison methods.
\vspace{-1mm}
\paragraph{Datasets and Setups.} Using the same synthetic data described in Sec.~\ref{sec:exp-online-tensor-factorization}, we randomly mask out $98\%$ of the entries and follow the same data preparation and mode growth settings. The results of mean curves are shown in Fig.~\ref{fig:synthetic}.
  
We also evaluate on two real-world datasets: (i) Indian Pines hyperspectral image dataset and (ii) a proprietary Covid disease counts data: location by disease by date, we call it the health tensor (CovidHT), and the results (refer to appendix~C.4) show that \method has the better fitness and $\mbox{\method}_{\light}$ has decent fitness with good efficiency.
%(ii) The Indian Pins hyperspectral image dataset contains 200 figures with size $145\times 145$ pixels for each. The sparsity of the first dataset is $95\%$ and the sparsity is $99\%$ for the second dataset. 
%Results are shown in Fig.~\ref{fig:synthetic} and Table~\ref{tb:stream-tensor-completion}.

%% file: sec6-conclusion.tex
\vspace{-1mm}
\section{Conclusion}
\vspace{-1mm}
This paper proposes a generalized online tensor factorization and completion framework, called \method, which can support various tensor evolving patterns and unifies existing online tensor formulations. Experiments confirmed that our \method can show good performance on the general setting, where most previous methods cannot support. Also, our \method provides comparable or better performance in each special setting. The \method package is currently open-sourced.

%% file: appendix.tex
\section{Matrix and Tensor Algebra, Notations} \label{sec:tensor-algebra-appendix}

\begin{table}[h!] \small \caption{Notations used in \method}\vspace{-3mm}
{\centering
\resizebox{0.49\textwidth}{!}{
\begin{tabular}{l|l} \toprule \textbf{Symbols} & \textbf{Descriptions} \\ 
	\midrule 
	$\odot$ & tensor Khatri-Rao product\\
	$\oast$ & tensor Hadamard product \\
	$\llbracket\cdot\rrbracket$ & tensor Kruskal product \\
	$R$ & tensor CP rank \\
	$\Omega^t\in\mathbb{R}^{I_1^t\times\cdots I_N^t}$ & tensor (observation) mask\\
	\multirow{2}{*}{$\tilde{\Omega}^t, \tilde{\Omega}^{t,old}$} & mask for new data and old data\\
	& where we have $\tilde{\Omega}^t \bigcup \tilde{\Omega}^{t,old} = \Omega^t$ \\
	$\mathcal{X}^t\in\mathbb{R}^{I_1^t\times\cdots I_N^t}$ & the underlying tensor\\
	$\mathcal{Y}^t\in\mathbb{R}^{I_1^t\times\cdots I_N^t}$ & low-rank approximation of $\mathcal{X}^t$\\
	$\mathbf{A}^{n,t}\in\mathbb{R}^{I_n^t\times R},~\forall n$ & the factor matrices\\
	$\mathbf{a}_{i_n}^{n,t}\in\mathbb{R}^{1\times R},~\forall n, i_n$ & the $i_n$-th row of the factor matrices \\
	$\mathbf{U}^{n,t}\in\mathbb{R}^{I_n^{t-1}\times R},~\forall n$ & the upper block matrices \\
	$\mathbf{L}^{n,t}\in\mathbb{R}^{(I_n^t-I_n^{t-1})\times R},~\forall n$ & the lower block matrices \\
	$\mathbf{P}_{i_n}^{n,t}\in\mathbb{R}^{R\times R}, \mathbf{q}_{i_n}^{n,t}\in\mathbb{R}^{1\times R}$ & auxiliary variables \\
	$\mathbf{P}_{\mathbf{U}}^{n,t}\in\mathbb{R}^{R\times R}, \mathbf{Q}_{\mathbf{U}}^{n,t}\in\mathbb{R}^{I_n^{t-1}\times R}$ & auxiliary variables for $\mathbf{U}^{n,t}$\\
	$\mathbf{P}_{\mathbf{L}}^{n,t}\in\mathbb{R}^{R\times R}, \mathbf{Q}_{\mathbf{L}}^{n,t}\in\mathbb{R}^{(I_n^t-I_n^{t-1})\times R}$ & auxiliary variables for $\mathbf{L}^{n,t}$\\
	\bottomrule
    \end{tabular}}
    \vspace{1mm}}
    {Note that (i) notation with superscript $^t$ means at time $t$; (ii) notations with $\hat{}$ means the initialized value or the estimated value.}
\label{tb:notations} \end{table}
	
\paragraph{Kronecker Product.} One important operation for matrices is the Kronecker product. For $\mathbf{A}\in\mathbb{R}^{I\times J}$ and $\mathbf{B}\in \mathbb{R}^{K\times L}$, their Kronecker product is defined by (each block is a scalar times matrix)
% \begin{equation} \small
%     \mathbf{A}\otimes\mathbf{B} = \begin{bmatrix}
%     \mathbf{a}_{11}\cdot \mathbf{B} & \mathbf{a}_{12}\cdot \mathbf{B} & \cdots&\mathbf{a}_{1J}\cdot \mathbf{B} \\
%     \mathbf{a}_{21}\cdot \mathbf{B}&\mathbf{a}_{22}\cdot \mathbf{B}&\cdots&\mathbf{a}_{2J}\cdot \mathbf{B}\\
%     \vdots&\vdots&\cdots&\vdots\\
%     \mathbf{a}_{I1}\cdot \mathbf{B}&\mathbf{a}_{I2}\cdot \mathbf{B}&\cdots&\mathbf{a}_{IJ}\cdot \mathbf{B}
%     \end{bmatrix}\in\mathbb{R}^{IK\times JL}. \notag
% \end{equation}
\begin{equation} \small
	\mathbf{A}\otimes\mathbf{B} = \begin{bmatrix}
		a_{11}\mathbf{B} & a_{12}\mathbf{B} & \cdots&a_{1J} \mathbf{B} \\
		a_{21}\mathbf{B}&a_{22}\mathbf{B}&\cdots &a_{2J}\mathbf{B}\\
		\vdots&\vdots&\cdots&\vdots\\
		a_{I1}\mathbf{B}&a_{I2}\mathbf{B}&\cdots&a_{IJ}\mathbf{B}
	\end{bmatrix}\in\mathbb{R}^{IK\times JL}. \notag
\end{equation}
\paragraph{Khatri–Rao Product.} 
Khatri-Rao product is another important product for matrices, specifically, for matrices with same number of columns. The Khatri-Rao product of $\mathbf{A}\in\mathbb{R}^{I\times J}$ and $\mathbf{B}\in \mathbb{R}^{K\times J}$ can be viewed as column-wise Kroncker product,
% \begin{equation}
%     \mathbf{A}\odot\mathbf{B} = \begin{bmatrix}
%     \mathbf{A}_{:1}\otimes\mathbf{B}_{:1},~ \mathbf{A}_{:2}\otimes\mathbf{B}_{:2},~ \cdots,~ \mathbf{A}_{:L}\otimes\mathbf{B}_{:L}
%     \end{bmatrix}, \notag
% \end{equation}
\begin{equation}
	\mathbf{A}\odot\mathbf{B} = \begin{bmatrix}
		\mathbf{a}^{(1)}\otimes\mathbf{b}^{(1)},\mathbf{a}^{(2)}\otimes\mathbf{b}^{(2)},\cdots,\mathbf{a}^{(L)}\otimes\mathbf{b}^{(L)}
	\end{bmatrix}, \notag
\end{equation}
where $\mathbf{a}^{(k)}$ and $\mathbf{b}^{(k)}$ are the $k$-th column of $\mathbf{A}$ and $\mathbf{B}$, and  $\mathbf{A}\odot\mathbf{B}\in\mathbb{R}^{IK\times L}$.

\paragraph{Tensor Unfolding.} This operation is to matricize a tensor along one mode. For tensor $\mathcal{X}\in\mathbb{R}^{I_1\times I_2\times I_3}$, we could unfold it along the first mode into a matrix $\mathbf{X}_{1}\in\mathbb{R}^{I_1\times I_2I_3}$. Specifically, each row of $\mathbf{X}_{1}$ is a vectorization of a slice in the original tensor; we have
\begin{equation*}
	\mathbf{X}_1(i,j\times I_3+k) = \mathcal{X}(i,j,k),~\forall i,j,k.
\end{equation*}
Here, to reduce clutter, we use Matlab-style notation to index the value. Similarly, for the unfolding operation along the second or third mode, we have
\begin{align*}
	\mathbf{X}_2(j,i\times I_3+k) &= \mathcal{X}(i,j,k)\in\mathbb{R}^{I_2\times I_1I_3},~\forall i,j,k,\\
	\mathbf{X}_3(k,i\times I_2+j) &= \mathcal{X}(i,j,k)\in\mathbb{R}^{I_3\times I_1I_2},~\forall i,j,k. 
\end{align*}

\paragraph{Hadamard Product.} The Hadamard product is the element-wise product for tensors of the same size. For example, the Hadamard product of two $3$-mode tensors $\mathcal{X},\mathcal{Y}\in\mathbb{R}^{I_1\times I_2\times I_3}$ is 
\begin{equation}
	\mathcal{Z} = \mathcal{X} \oast \mathcal{Y} \in\mathbb{R}^{I_1\times I_2\times I_3}. \notag
\end{equation}

\paragraph{Canonical Polyadic (CP) Factorization.} One of the common compression methods for tensors is CP factorization \cite{hitchcock1927expression,carroll1970analysis}, also called CANDECOMP/PARAFAC (CP) factorization, which represents a tensor by multiple rank-one components. For example, let $\mathcal{X}\in\mathbb{R}^{I_1\times I_2\times I_3}$ be an arbitrary $3$-order tensor of CP-rank $R$, then it can be expressed exactly by factor matrices $\mathbf{U}\in\mathbb{R}^{I_1\times R},\mathbf{V}\in\mathbb{R}^{I_2\times R},\mathbf{W}\in\mathbb{R}^{I_3\times R}$ as
\begin{equation*}
	\mathcal{X} = \sum_{r=1}^R\mathbf{u}^{(r)}\circ\mathbf{v}^{(r)}\circ \mathbf{w}^{(r)} = \llbracket\mathbf{U},\mathbf{V},\mathbf{W}\rrbracket,
\end{equation*}
where $\circ$ is vector outer product, $\mathbf{u}^{(r)},\mathbf{v}^{(r)},\mathbf{w}^{(r)}$ are the $r$-th column vectors.

\section{Complexity Analysis} \label{sec:complexity-analysis}

Follow the notations in Section~2 of main paper, we analyze the space and time complexity of the optimization algorithm. Assume the CP-rank $R$ satisfies: $I_n^t\gg R,~\forall t,~\forall n$. Let $S^t=\sum_{n=1}^NI_n^t$, $P^t=\prod_{n=1}^NI_n^t$, and $nnz(\cdot)$ indicates the number of non-zero entries. 

\paragraph{Space Complexity.} The space complexity records the storage needed for input at time $t$. For objective $\mathcal{L}$, we need the previous factors, $S^{t-1}R$, and the current data, $nnz(\Omega^t)$ (note that, nonzero entries of the mask indicates the observed data). In total, $S^{t-1}R+nnz(\Omega^t)$ is the input size; for the efficient version $\mathcal{L}_{\light}$, the old unchanged data is not needed. Therefore, the input size is $S^{t-1}R+nnz(\tilde{\Omega}^t)$.

\vspace{-1mm}
\paragraph{Time Complexity.} For time complexity, the analysis is under the sparse or dense strategies in Sec 3.2 of the main paper.
\begin{itemize}
    \vspace{-0.5mm}
	\item {\em Sparse strategy.} In Eqn.~(10), the overall cost of computing $\mathbf{P}^{n,t}_{i_n}$ is $\mathcal{O}(nnz(\Omega^t)NR(N+R))$; In Eqn.~(11), the overall cost of computing $\mathbf{q}^{n,t}_{i_n}$ is $\mathcal{O}(nnz(\Omega^t)N^2R)$; the cost of matrix inverse for all $\mathbf{P}^{n,t}_{i_n}$ is $\mathcal{O}(S^tR^3)$ by Cholesky decomposition (which could be omitted since $nnz(\Omega^t)\gg S^t$). Overall, the time complexity for $\mathcal{L}$ is $\mathcal{O}(nnz(\Omega^t)NR(N+R))$. The time complexity for $\mathcal{L}_{\light}$ is $\mathcal{O}(nnz(\tilde{\Omega}^t)NR(N+R))$.
	\vspace{-0.5mm}
	\item {\em Dense strategy.} First, the cost of imputing the tensor is $\mathcal{O}(P^tR)$. Then, in Eqn.~(13), the overall cost of MTTKRP ($\mathbf{Q}^{n,t}_{\mathbf{U}}$ and $\mathbf{Q}^{n,t}_{\mathbf{L}}$) is $\mathcal{O}(NP^tR)$; the cost of computing $\mathbf{P}^{n,t}_{\mathbf{U}}$ and $\mathbf{P}^{n,t}_{\mathbf{L}}$ is $\mathcal{O}(S^tR^2)$ and the cost of calculating their inverse is $\mathcal{O}(NR^3)$ (these two quantity could be omitted since $P^t\gg S^t, P^t\gg R^2$).
	The dominant complexity are imputation and MTTKRP, which sum up to $\mathcal{O}(NP^tR)$ for $\mathcal{L}$. The cost of $\mathcal{L}_{\light}$ is $\mathcal{O}(NP^tR/I_n^t)$ with only mode growth (one new slice at a time), but it can be up to $\mathcal{O}(NP^tR)$ in more complex scenarios.
\end{itemize}

For a comprehensive comparison, we listed the space and time complexity of all baselines in Table~\ref{tb:complexity}, while for different settings, the baselines are different.

\begin{table}[tbp!] 
	\resizebox{.47\textwidth}{!}{\begin{tabular}{lcc} \toprule \textbf{Model} & \textbf{Space Complexity} & \textbf{Time Complexity} \\ \midrule
			\multicolumn{3}{l}{\bf For Sec.~5.2 (the general setting) and Sec.~5.4 (the completion setting):} \\
			EM-ALS & $nnz(\Omega^t)+(\bar{S}+I_N)R$ & $\mathcal{O}(NR\bar{P}I_N)$\\
			CPC-ALS & $nnz(\Omega^t)+(\bar{S}+I_N)R$ & $\mathcal{O}(nnz(\Omega^t)NR(N+R))$\\
% 			CPC-ALS-REF & $nnz(\Omega^t)$ & $\mathcal{O}(nnz(\Omega^t)NR(N+R))$\\
			OnlineSGD & $nnz(\tilde{\Omega}^t)+(\bar{S}+I_N)R$ & $\mathcal{O}(nnz(\tilde{\Omega}^t)NR(N+R))$\\
			OLSTEC & $nnz(\tilde{\Omega}^t)+(\bar{S}R+2\bar{S}+I_N)R$ & $\mathcal{O}(nnz(\tilde{\Omega}^t)NR(N+R))$ \\
			$\mbox{\method}_{\light}$ & $nnz(\tilde{\Omega}^t)+(\bar{S}+I_N)R$ & $\mathcal{O}(nnz(\tilde{\Omega}^t)NR(N+R))$\\
			\method & $nnz(\Omega^t)+(\bar{S}+I_N)R$ & $\mathcal{O}(nnz(\Omega^t)NR(N+R))$\\
			\midrule
			\multicolumn{3}{l}{\bf For Sec.~5.3 (the factorization setting):} \\
			OnlineCPD & $\bar{P}+(2\bar{S}+I_N)R+(N-1)R^2$ & $\mathcal{O}(NR\bar{P})$\\
			MAST & $\bar{P}+3(\bar{S}+I_N)R$ & $\mathcal{O}(NR\bar{P})$\\
%			SDT & $\bar{P}+(\bar{S}+\bar{P}+2I_N)R+3R^2$ & $\mathcal{O}(R^2(\bar{P}+I_N))$\\
			RLST & $\bar{P}+(\bar{S}+2\bar{P}+I_N)R+2R^2$ & $\mathcal{O}(R^2\bar{P})$ \\
			CPStream & $\bar{P}+(\bar{S}+I_N+R)R$ & $\mathcal{O}(NR\bar{P})$\\
			$\mbox{\method}_{\light}$ & $\bar{P}+(\bar{S}+I_N)R$ & $\mathcal{O}(NR\bar{P})$\\
			\method & $\bar{P}I_N+(\bar{S}+I_N)R$ & $\mathcal{O}(NR\bar{P}I_N)$\\
%			\midrule
%			{\bf For Sec.~\ref{sec:exp-stream-tensor-completion}:} \\
%			EM-ALS & $nnz(\Omega^t)+(\bar{S}+I_N)R$ & $\mathcal{O}(N\bar{P}I_NR)$\\
%			$\mbox{\method}_{\light}$ & $nnz(\tilde{\Omega}^t)+(\bar{S}+I_N)R$ & $\mathcal{O}(nnz(\tilde{\Omega}^t)NR(N+R))$ \\
%			\method & $nnz(\Omega^t)+(\bar{S}+I_N)R$ & $\mathcal{O}(nnz(\Omega^t)NR(N+R))$\\
		\bottomrule \end{tabular}}
		\vspace{-1mm}
		\caption{Complexity per iteration at Time $t$, where $\bar{P}=P/I_N,\bar{S}=S-I_N$ (refer to Sec.~\ref{sec:complexity-analysis}) and we remove the superscript $^t$ whenever there is no ambiguity.}
		\vspace{-1mm}
		\label{tb:complexity}
	%\noindent~ * For convenience, we define $\bar{P}=P/I_N,\bar{S}=S-I_N$ (refer to Sec.~\ref{sec:complexity-analysis}) and remove the superscript $^t$ if there is not ambiguity. Some results are borrowed from \cite{zhou2018online}. For both settings in Sec.~\ref{sec:exp-online-tensor-factorization} and Sec.~\ref{sec:exp-stream-tensor-completion}, only one new slide is added at each time step. $\tilde{\Omega}^t$ means the set of newly added/updated elements while $\Omega^t$ indicates the current overall elements at time $t$. Note that, the table only shows the time complexity for one iteration, while some methods takes several iterations at one time step, e.g., CPC-ALS-S, MAST and CPStream.
	 \end{table}

\section{Additions for Experiments} \label{sec:additional-experiment}

All experiments use $R=5$ and are implemented by {\em Python 3.8.5}, {\em Numpy 1.19.1} and the experiments are conducted on a Linux workstation with 256 GB memory, 32 core CPUs (3.70 GHz, 128 MB cache)  In our experiments of the main text, we use the sparse strategy for the general case and the tensor completion case and use dense strategy for the factorization case.

{\bf Average PoF (Avg. PoF).} Note that, in section 5.2 of the main paper, the {\em Avg. PoF} is calculated by averaging the PoF scores over the time steps (remember that we calculate a PoF score per time step), and the mean and standard deviation values are calculated on five runs with different random seeds. 
\subsection{Datasets Description} \label{sec:dataset_description}
\begin{itemize}
    \item {\bf JHU Covid Dataset.} As mentioned in the main text, this dataset\footnote{https://github.com/CSSEGISandData/COVID-19} \cite{dong2020interactive} was collected from Apr. 6, 2020 to Oct. 31, 2020. We model the data as a fourth-order tensor and treat 3 categories: new cases, deaths, and hospitalization as the feature dimension. The collected data covers 51 states across the US. This data contains 209 generation dates (GD) and 8 loading dates (LD) for each data point. The leading $50\%$ slices (before Jul. 16) are used as preparation data to obtain the initial factors. The reflection of three evolving patterns: at each timestamp, the GD will increase by 1 and the LD dimension will be filled accord to how previous data is received. Later, we add random perturbations to randomly selected $2\%$ eisting data with value changes uniformly of $[-5\%,5\%]$ at each time step.
    
	\item {\bf Patient Claim Data.} This dataset can also be modeled as a fourth-order tensor, which contains 56 states, 22 disease categories, 10 LD's (for the delayed updates), and 104 GD's (104 weeks over the year 2018 and 2019, there is one generated service data point for each week). The leading $50\%$ slices (data of year 2018) are used as preparation data to obtain the initial factors. The reflection of three evolving patterns: at each timestamp, the GD will increase by 1 and the LD dimension will be filled accordingly. Later, at each time step, we add random perturbations to randomly selected $2\%$ eisting data with value changes uniformly of $[-5\%,5\%]$.
	
	\smallskip
	For the above two datasets, the GD mode is the growing mode (referring to change type (i)), and the LD (delayed updates) causes the incompleteness of previous tensor slices, while the incomplete slices will be filled in later dates (referring to change type (ii)). By fixing the first and the second modes, we show a diagram (figure below) of how the GD and LD evolve. At each time step, a new GD is generated while the LDs for the previous GD's can be filled (assuming that all GD's have the same number of LDs in total). A special setting is discussed in a recent paper \cite{qian2021multi}.
	
	\begin{figure}[htbp!] \centering
	\includegraphics[width=3.1in]{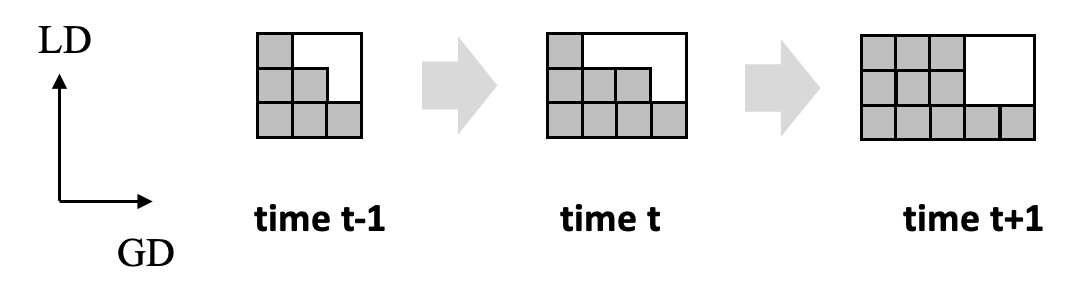} \vspace{-2mm}
    \end{figure}
    
    Additionally, to simulate change type (iii) on these two datasets, we add random perturbations to the existing tensor elements in the experiments.

	\item {\bf FACE-3D.} ORL Database of Faces\footnote{https://cam-orl.co.uk/facedatabase.html} contains 400 shots of face images with size 112 pixels by 99 pixels. To construct a third-order tensor, we treat each shot as a slice and concatenate them together. The first $10\%$ leading shots are used as preparation data.
	\item {\bf GCSS.} This is a public dataset\footnote{https://pair-code.github.io/covid19\_symptom\_dataset/} containing google covid-19 symptom search results during the complete year 2020. The data can be viewed as a third-order tensor: 50 states, 422 keywords, and 362 days. The first $10\%$ leading dates are used as preparation data.
	\item {\bf Indian Pines.} This is also an open dataset\footnote{https://purr.purdue.edu/publications/1947/1} containing 200 shots of hyperspectral images with size 145 pixels by 145 pixels. The data is also modeled as a third-order tensor. We manually mask out $98\%$ of the tensor elements for the online tensor completion problem. Later, the masked out entries are used as ground truth to evaluate our completion results. The first $10\%$ leading images are used as preparation data.
	\item {\bf CovidHT.} This is a proprietary covid-19 related disease tensor with 420 zip codes, 189 disease categories, and 128 dates. We model it as a third-order tensor. Similar to the Indian Pines datasets, we manually mask out $98\%$ of the tensor elements for later evaluation, and the first $10\%$ leading shots are used as preparation data.
\end{itemize}

\subsection{Hyperparameter settings}
Our framework has two hyperparameters, the first one is $\beta$, which is the weight for L2 regularization. We set $\beta=10^{-5}$ for all experiments and it does not change w.r.t. time step $t$. Another hyperparameter is $\alpha$, controlling the importance of historical reconstruction. Let us set $I^t$ to be the dimension of the growing mode. We listed the $\alpha$ used in the experiments in Table~\ref{tb:hyperparameters}. There are several advices on choosing the hyperparameters: (i) for any settings included with real-world datasets, the $\alpha$ can be chosen based on the simulation results of the synthetic data; (ii) we keep a larger $\alpha$ for the real-world data (in factorization case), since the real data is usually not smooth and the statistics from old slices to new slices could change. A larger $\alpha$ would make transition from time $(t-1)$ to time $t$ more smooth, which makes the final results more stable; (iii) compared to \method, our $\mbox{\method}_{\light}$ does not keep the historical data, so it relies on the historical reconstruction more, then we set a larger $\alpha$, which is 100 times the $\alpha$ for \method. The details for different settings can refer to: the general case (Sec. 5.2 of main text), factorization case (Sec. 5.3 of main text), completion cases (Sec. 5.4 of main text).
%First, for the general case, we set $\alpha=0.02/I^t$ for $\mbox{\method}_{\light}$ and $\alpha=0.02/I^t$ for \method. For the special factorization case, we set $\alpha=2/I^t$ for \method on three datasets; for $\mbox{\method}_{\light}$, we set $\alpha=2/I^t$ for the synthetic data and $\alpha=min(1,200/I^t)$ for the real data. For the special completion case, we set $\alpha=0.5/I^t$ for both $\mbox{\method}_{\light}$ and \method on all datasets. The choice of $\alpha$ depends on 

\begin{table}[h!] 
	\resizebox{0.47\textwidth}{!}{\begin{tabular}{c|c|c} \toprule 
			{\bf Settings} & \method & $\mbox{\method}_{\light}$ \\
			\midrule 
			general case & $\alpha=0.02/I^t$ &  $\alpha=2/I^t$ \\
			\midrule
			\multirow{2}{*}{factorization case} & $\alpha=0.02/I^t$ for synthetic data &  $\alpha=2/I^t$ for synthetic data \\
			& $\alpha=2/I^t$ for real data &  $\alpha=min(1,200/I^t)$ for real data \\ 
			\midrule 
			{completion case} & $\alpha=0.005/I^t$  &  $\alpha=0.5/I^t$ \\
% 			\midrule 
% 			other case 1 & $\alpha=0.02/I^t$ &  $\alpha=2/I^t$ \\
% 			\midrule 
% 			other case 2 & $\alpha=0.02/I^t$ &  $\alpha=2/I^t$ \\
			\bottomrule
	\end{tabular}}
	\caption{Hyperparameter $\alpha$ Settings for All Experiments}
	\vspace{-1mm} \label{tb:hyperparameters}
\end{table}

%For $\mathcal{L}_{\light}$, we set $\alpha=2/I^t$ for the general case and tsynthetic data and $\alpha=min(0.1,\frac{20\times nnz(\tilde{\Omega}^t)}{nnz(\Omega^t)})$ for real-world data. For $\mathcal{L}_{\light}$, we let $\alpha=2/I^t$ and $\alpha=min(0.1,20/I^t)$ for synthetic and real-world data respectively, where $I^t$ is the size of the growing mode at time $t$. We keep a larger $\alpha$ for real-world data, since the real data is usually not smooth and the statistics from old slices to new slices could change. A larger $\alpha$ would make transition from time $(t-1)$ to time $t$ more smooth, which makes the final results more stable.

\subsection{Exp. for Online Tensor Factorization} \label{app:online-tensor-factorization}
The setting is discussed in the main paper, however, we move the result table here due to space limitation.

{Note that the contribution of our model is to (a) provide a unified framework and can handle
both online tensor factorization and completion (previous models are designed only for either); (b) show that our model can provide comparable or better performance over all baselines in all settings. Here, section C.3 and C.4 are to empirically show (b).}

\begin{table}[h!] 
		\resizebox{0.47\textwidth}{!}{\begin{tabular}{ccccc} \toprule 
				\multirow{2}{*}{\bf Model} & \multicolumn{2}{c}{\bf FACE-3D} & \multicolumn{2}{c}{\bf GCSS} \\
				\cmidrule{2-5}
				& Total Time (s) & Avg. PoF & Total Time (s) & Avg. PoF \\
				\midrule
				OnlineCPD & 13.61 $\pm$ 0.040 & 0.7447 $\pm$ 1.129e-3 & 26.38 $\pm$ 0.035 & 0.9258 $\pm$ 3.053e-4\\
                MAST & 98.52 $\pm$ 0.012 & 0.7464 $\pm$ 1.405e-3 & 159.21 $\pm$ 0.192 & 0.9278 $\pm$ 2.437e-4\\
                RLST & 13.33 $\pm$ 0.033 & 0.7216 $\pm$ 2.342e-3 & 26.64 $\pm$ 0.004 & 0.6725 $\pm$ 3.653e-2\\
                CPStream & 18.35 $\pm$ 0.005 & 0.7446 $\pm$ 1.281e-3 & 43.61 $\pm$ 0.112 & 0.9256 $\pm$ 5.192e-4\\
                $\mbox{\method}_{\light}$ & 13.77 $\pm$ 0.165 & 0.7452 $\pm$ 1.208e-3 & 26.75 $\pm$ 0.094 & 0.9270 $\pm$ 1.022e-4\\
                \method & 17.14 $\pm$ 0.071 & 0.7473 $\pm$ 1.315e-3 & 34.17 $\pm$ 0.338 &  0.9297 $\pm$ 3.203e-4\\
				\bottomrule
	\end{tabular}}
	\caption{Results for Online Tensor Factorization}
	\vspace{-1mm} \label{tb:online-tensor-factorization}
	\vspace{-2mm}
\end{table}

\paragraph{Result Analysis.} On the real data, OnlineCPD and our $\mbox{\method}_{\light}$ show great fitness and the best efficiency. Though \method presents the best fitness score, it is relatively slower than the best model. The baselines MAST and CPStream are expensive since they use ADMM to iteratively update the factors within each time step, which is time-consuming.

\subsection{Exp. for Online Tensor Completion}  \label{app:stream-tensor-completion}
Due to space limitation, we move the experimental results here and give result summary in the main text.
\begin{table}[h!] 
	\resizebox{0.47\textwidth}{!}{\begin{tabular}{ccccc} \toprule 
			\multirow{2}{*}{\bf Model} & \multicolumn{2}{c}{\bf Indian Pines} & \multicolumn{2}{c}{\bf CovidHT} \\
			\cmidrule{2-5}
			& Total Time (s) & Avg. PoF & Total Time (s) & Avg. PoF \\
			\midrule
			EM-ALS & 17.15 $\pm$ 0.022 & 0.8839 $\pm$ 5.696e-4 & 25.71 $\pm$ 0.148 & 0.5432 $\pm$ 2.317e-2\\
			EM-ALS (decay) & 17.05 $\pm$ 0.014 & 0.8845 $\pm$ 5.452e-4 & 25.69 $\pm$ 0.115 & 0.5463 $\pm$ 2.314e-2\\
			OnlineSGD & 11.33 $\pm$ 0.025 & 0.8576 $\pm$ 3.502e-4 & 15.99 $\pm$ 0.061 & 0.4422 $\pm$ 1.185e-2\\
			OLSTEC & 11.32 $\pm$ 0.014 & 0.8864 $\pm$ 7.447e-5 & 16.05 $\pm$ 0.141 & 0.6503 $\pm$ 7.041e-3\\
			$\mbox{\method}_{\light}$ & 11.82 $\pm$ 0.031 & 0.8923 $\pm$ 5.083e-4 & 16.44 $\pm$ 0.039 & 0.6612 $\pm$ 1.321e-2\\
			\method & 19.89 $\pm$ 0.545 & 0.8970 $\pm$ 1.281e-3 & 27.92 $\pm$ 0.091 & 0.6792 $\pm$ 2.602e-2\\
			\bottomrule
	\end{tabular}}
\caption{Results for Online Tensor Completion}
\vspace{-1mm} \label{tb:stream-tensor-completion}
\end{table}

\paragraph{Result Analysis.} On the real-world data, baseline OLSTEC and our $\mbox{\method}_{\light}$ show decent performance, and they both outperform other baseline methods in terms of both fitness and speed, though \method has the best fitness score consistently.

\subsection{Discussion on Different Evolving Patterns} Here, we summarize some intuitions and conclusions for the evolving patterns.
\begin{itemize}
    \item Mode growth may lead to better or poor PoF based on data
quality in new dimensions. With more filled missing values,
PoF will intuitively increase. These two patterns will
increase the running time because the data size (or dimension
size) increases.
\item According to Table 2 of the main paper, with the additional value
update pattern (perturbed version), the running time of all
models increases slightly while the final PoF may be improved
or not (based on how good the updated value is).
\end{itemize}

\subsection{Ablation Study on Mask Density} \label{sec:ablation-study}
\begin{figure}[htbp!] \centering
    \vspace{-2mm}
	\includegraphics[width=3.1in]{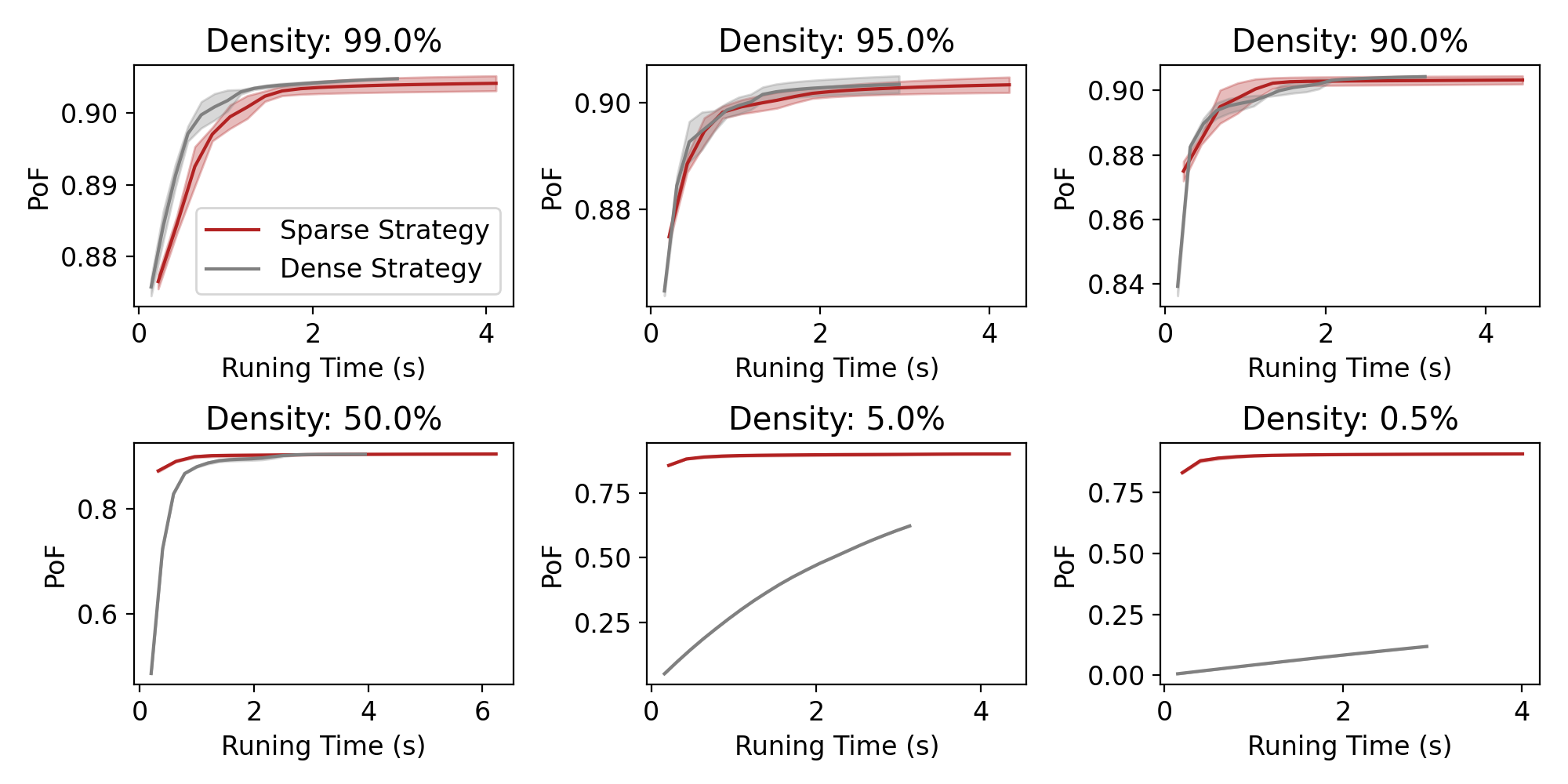} \vspace{-2mm}\caption{Comparison with Different Density on Mask} \vspace{-2mm}\label{fig:masked_vs_imputed} 
\end{figure}

% This section evaluates the performance of sparse and dense strategies in Sec.~\ref{sec:rw-als}. We construct low-rank tensors with size $(I_1,I_2,I_3,R)=(100,100,100,5)$. The tensors are fixed with six levels of density, and both strategies run 100 iterations. We plot the running time versus PoF in Fig.~\ref{fig:masked_vs_imputed}.
This section evaluates the performance of sparse and dense strategies in Sec.~3.2. We experiment on Indian Pines dataset with the full tensors and generate random masks with six density levels. Both strategies run for 25 iterations. We plot the running time versus PoF metric in Fig.~\ref{fig:masked_vs_imputed}.

\paragraph{Result Analysis.} We observe that dense strategy works well and more efficiently on masks with high density while the sparse strategy is more advantageous when the density is low. {\em Thus, we use the sparse strategy for the general case in Sec.~5.2 and completion case in Sec.~5.4, and use the dense strategy (without the imputation step) for common factorization setting in Sec.~5.3.} Note that, our models run either strategy for one iteration per time step.